\documentclass[sigconf]{acmart}

\usepackage{multirow}
\usepackage{diagbox}
\usepackage{subfigure}
\usepackage{dsfont}

\usepackage{balance}

\renewcommand{\shortauthors}{Hongbin Xu et al.}

\AtBeginDocument{%
  }

\copyrightyear{2023}
\acmYear{2023}
\setcopyright{acmlicensed}\acmConference[MM '23]{Proceedings of the 31st ACM International Conference on Multimedia}{October 29-November 3, 2023}{Ottawa, ON, Canada}
\acmBooktitle{Proceedings of the 31st ACM International Conference on Multimedia (MM '23), October 29-November 3, 2023, Ottawa, ON, Canada}
\acmPrice{15.00}
\acmDOI{10.1145/3581783.3611931}
\acmISBN{979-8-4007-0108-5/23/10}

\begin{document}

\title{Semi-supervised Deep Multi-view Stereo}

\author{Hongbin Xu}
\orcid{0000-0002-3455-1527}
\authornote{Both authors contributed equally to this work.}
\affiliation{
    \institution{South China University of Technology}
    \city{Guangzhou}
    \country{China}
}
\email{hongbinxu1013@gmail.com}

\author{Weitao Chen\footnotemark[1]}
\orcid{0000-0003-1796-2671}
\affiliation{
    \institution{Alibaba Group}
    \city{Hangzhou}
    \country{China}
}
\email{hillskyxm@gmail.com}

\author{Yang Liu}
\orcid{0000-0001-8035-4383}
\affiliation{
    \institution{Alibaba Group}
    \city{Hangzhou}
    \country{China}
}

\author{Zhipeng Zhou}
\orcid{0009-0007-7988-8856}
\affiliation{
    \institution{Alibaba Group}
    \city{Hangzhou}
    \country{China}
}

\author{Haihong Xiao}
\orcid{0000-0002-3543-9262}
\affiliation{
    \institution{South China University of Technology}
    \city{Guangzhou}
    \country{China}
}

\author{Baigui Sun}
\orcid{0000-0001-7722-4748}
\affiliation{
    \institution{Alibaba Group}
    \city{Hangzhou}
    \country{China}
}

\author{Xuansong Xie}
\orcid{0000-0002-3671-799X}
\affiliation{
    \institution{Alibaba Group}
    \city{Hangzhou}
    \country{China}
}

\author{Wenxiong Kang}
\authornote{Corresponding author.}
\orcid{0000-0001-9023-7252}
\affiliation{
    \institution{South China University of Technology}
    \city{Guangzhou}
    \country{China}
}
\additionalaffiliation{
    \institution{Pazhou Laboratory}
    \city{Guangzhou}
    \country{China}
}
\email{auwxkang@scut.edu.cn}

\renewcommand{\shortauthors}{Hongbin Xu et al.}

\begin{abstract}
Significant progress has been witnessed in learning-based Multi-view Stereo (MVS) under supervised and unsupervised settings. 
To combine their respective merits in accuracy and completeness, meantime reducing the demand for expensive labeled data, this paper explores the problem of learning-based MVS in a semi-supervised setting that only a tiny part of the MVS data is attached with dense depth ground truth. 
However, due to huge variation of scenarios and flexible settings in views, it may break the basic assumption in classic semi-supervised learning, that unlabeled data and labeled data share the same label space and data distribution, named as semi-supervised distribution-gap ambiguity in the MVS problem. 
To handle these issues, we propose a novel semi-supervised distribution-augmented MVS framework, namely SDA-MVS. 
For the simple case that the basic assumption works in MVS data, consistency regularization encourages the model predictions to be consistent
between original sample and randomly augmented sample. 
For further troublesome case that the basic assumption is conflicted in MVS data, we propose a novel style consistency loss to alleviate the negative effect caused by the distribution gap. 
The visual style of unlabeled sample is transferred to labeled sample to shrink the gap, and the model prediction of generated sample is further supervised with the label in original labeled sample.
The experimental results in semi-supervised settings of multiple MVS datasets show the superior performance of the proposed method.
With the same settings in backbone network, our proposed SDA-MVS\footnote{The code is released in: \url{https://github.com/ToughStoneX/Semi-MVS}.} outperforms its fully-supervised and unsupervised baselines.
\end{abstract}






\maketitle

\section{Introduction}
\label{sec:intro}

\begin{figure*}
\centering
\subfigure[Visualization of the confusion matrix measuring distribution gap (MMD distance) among different scenes in DTU, BlendedMVS, and GTA-SFM datasets.]{
\includegraphics[width=.9\textwidth]{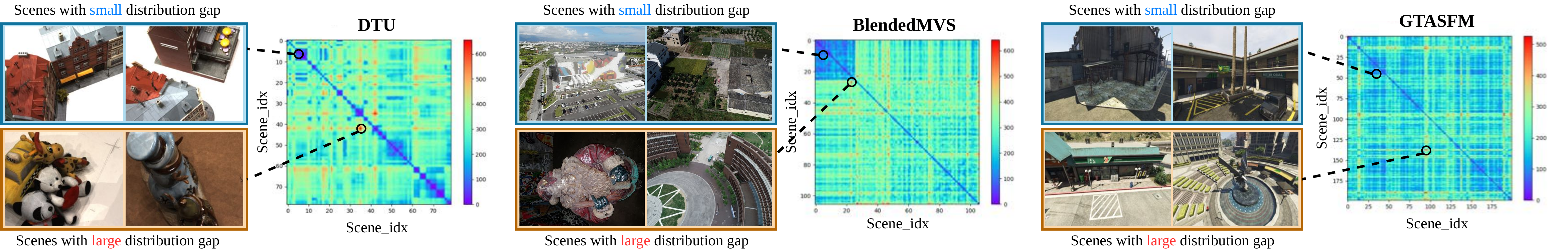}
}
\quad
\subfigure[Geometry-dropping problem when directly applying neural style transfer algorithms.
We visualize 3D consistency via running a MVS algorithm (COLMAP {\cite{schonberger2016pixelwise}}).
The geometric details may be lost after style transfer (2-nd row), compared with the original images (1-st row).
After being post-processed by GPM, the geometric details under style transfer can be preserved completely (3-rd row).]{
\includegraphics[width=.9\textwidth]{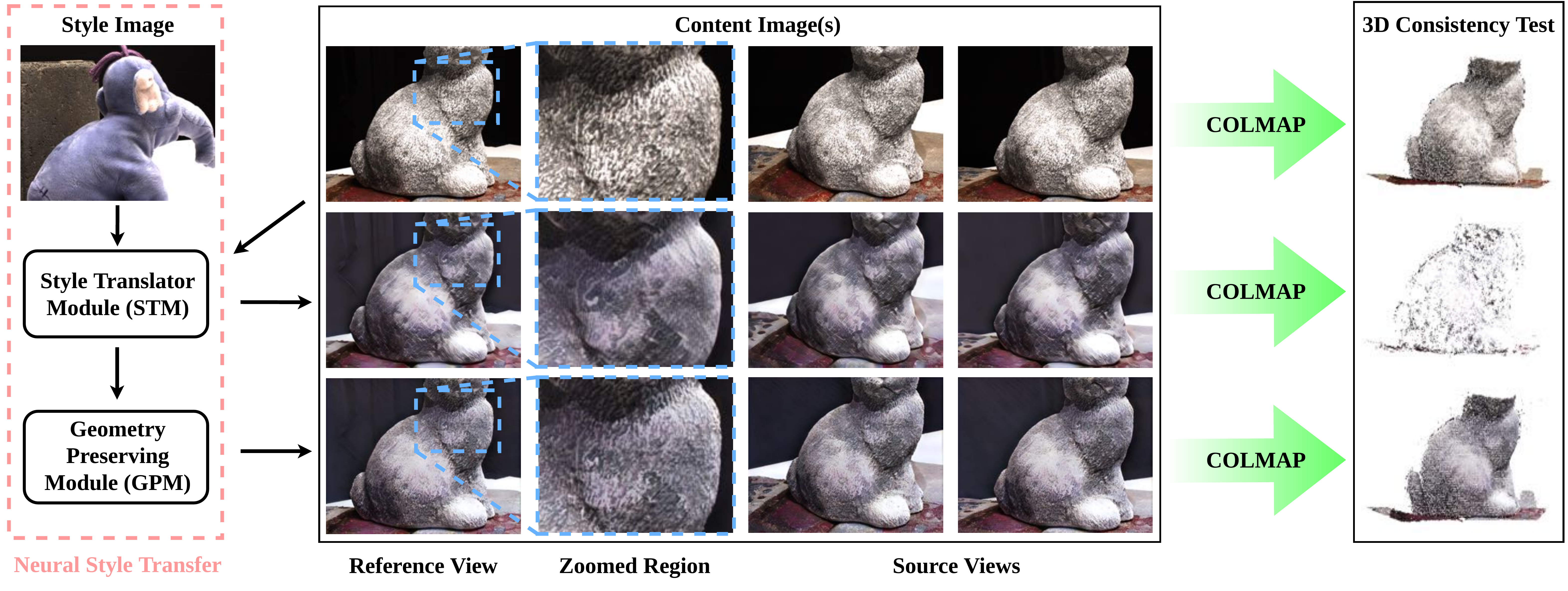}
}
\caption{Visualization proofs of the semi-supervised MVS problem.}
\label{fig:motivation}
\end{figure*}

Multi-view Stereo (MVS) is one of the cornerstone problems in computer vision, which reconstructs dense 3D geometry from calibrated multi-view images.
Stereoscopic vision for 3D reconstruction is on the cusp of many industrial applications such as autonomous driving, robotics, and virtual reality for decades.
Recent MVS works {\cite{yao2018mvsnet,yao2019recurrent,xiaodong2020cascade}} extend the traditional approaches to deep-learning based methods, and improve the 3D reconstruction performance with the blessing of large-scale MVS datasets {\cite{jensen2014large,knapitsch2017tanks}}.
Despite their ideal performance, there have been non-negligible difficulties in collecting dense 3D ground truth annotations, which may hamper the generalization to new domains.
Specifically, collecting accurate and complete 3D ground truth {\cite{jensen2014large,knapitsch2017tanks}} requires tedious collection process with a fixed active sensor, as well as labor-intensive post-processing procedures to remove outliers like moving objects in a static scene.
Thus, unsupervised/self-supervised MVS methods are proposed to avoid the dependence on the expensive 3D ground truth, which build the depth estimation problem as an image reconstruction problem with photometric consistency {\cite{khot2019learning,huang2021m3vsnet,xu2021self,xu2021digging}}.
With the help of these methods, the perplexity of 3D annotations can be relieved, meantime achieving amazing 3D reconstruction quality {\cite{xu2021self}}.

Rethinking the merits and demerits of unsupervised and supervised MVS compared with each other, we can have the following findings: 
1) \emph{Considering 3D reconstruction completeness, unsupervised MVS performs better than supervised MVS.}
Since the self-supervision loss built on photometric consistency excavates supervision signals on all available pixels in the image, unsupervised MVS has \emph{more complete} regions with valid supervision constraints compared with supervised MVS which only has limited label-intensive annotations.
2) \emph{Considering 3D reconstruction accuracy, supervised MVS performs better than unsupervised MVS.}
Different from the valid supervision in supervised MVS, the dense self-supervision loss is usually \emph{not accurate enough}, because it may be invalid on many unexpected cases, such as color constancy ambiguity {\cite{xu2021self}}, textureless backgrounds {\cite{xu2021digging}} and occluded regions {\cite{dai2019mvs2}}.

Instead of merely staring at the demerits of unsupervised and supervised MVS methods for improvements, we can see that they are \emph{complementary to each other} on their respective merits of improving completeness and accuracy.
In this paper, to combine the merits of unsupervised and supervised MVS, we explore a novel \emph{semi-supervised MVS} problem, which assumes that only a tiny part of the MVS dataset has 3D annotations.
Specifically, it has an intractable risk of breaking the basic assumption in the standard semi-supervised classification problem {\cite{xie2020unsupervised,miyato2018virtual,grandvalet2004semi}}, that \emph{labeled and unlabeled data come from the same label space, following independently identical distribution(i.i.d.)}. 
As shown in Fig. \ref{fig:motivation} (a)), the inherent distribution gap among different scenes in MVS problem may confuse the learning process, namely \emph{semi-supervised distribution-gap ambiguity}.



To handle the problems, we propose a novel MVS framework, called semi-supervised distribution-augmented MVS framework (SDA-MVS).
1) The basic framework of SDA-MVS handles the labeled samples and unlabeled samples differently.
The labeled samples are supervised under the common regime of supervision loss {\cite{yao2018mvsnet}} measuring the difference between the prediction and ground truth.
The basic photometric consistency loss {\cite{khot2019learning}} is used to supervise the unlabeled samples.
No extra extensions {\cite{xu2021self,xu2021digging,dai2019mvs2}} of the self-supervision loss are used to maintain a concise pipeline.
2) For the simple case that \emph{the assumption works}, consistency regularization loss is used to minimize the difference of depth predictions with or without random data-augmentation.
Following the low-density assumption {\cite{grandvalet2004semi}}, the low-density separation boundary among classes is enforced through the invariance against data-augmentations and proximity in latent space, meantime spreading the priors from labeled data to unlabeled data.
3) For further troublesome case that \emph{the assumption fails}, we propose a style consistency loss consisting of a style translation module (STM) and geometry-preserving module (GPM).
Taking inspiration from neural style transfer algorithms {\cite{gatys2015neural,li2017universal}}, STM transfers the visual styles from unlabeled MVS images to labeled MVS images.
However, the style transfer algorithms may bring unexpected distortions in the generated images, which may corrupt the cross-view correspondence relationship in the MVS data (further discussed in Fig. \ref{fig:motivation} (b)). 
Consequently, GPM utilizes a spatial propagation network {\cite{li2017universal}} to regularize the affinity of images, acting as an anti-distortion module towards unexpected distortions.
The ground truth is then used to supervise the generated MVS images after style translation, diminishing the negative effect of distribution discrepancy between labeled and unlabeled MVS data.

In summary, our contributions are listed as follows:
1) In a semi-supervised setting which assumes only a small part of the MVS dataset are labeled, we firstly investigate a novel semi-supervised distribution-perturbed problem and propose a novel MVS framework named SDA-MVS;
2) To handle the natural distribution gap between labeled and unlabeled MVS data, we propose a style consistency loss to alleviate the problem.
3) For evaluation, the experimental results on DTU, BlendedMVS, GTA-SFM, and Tanks\&Temples demonstrate the superior performance of the proposed method.
We further extend the semi-supervised setting to the semi-supervised domain adaptation task in multiple MVS datasets and evaluate the effectiveness of SDA-MVS.



\section{Related Work}
\label{sec:related}


\subsection{Fully-supervised Multi-view Stereo}
\label{sec:related:full_sup_mvs}

Thanks to the bless of deep neural networks, learning-based methods have been successfully developed on MVS reconstruction.
The pioneering work of MVSNet {\cite{yao2018mvsnet}} proposes an end-to-end network for multi-view depth estimation.
The multi-view feature maps are extracted with a shallow Convolutional Neural Network (CNN) with shared weights.
Then, the cross-view feautre maps are projected to the reference frustum via differentiable homography warping to construct a cost volume following the regime of plane sweeping {\cite{nievergelt1982plane}}.
On each hypothetical depth plane, each element of cost volume represents the similarity score of the matching points among views.
After regularizing by 3D CNN, the predicted probability volume is used to regress depth map via soft-argmin {\cite{kendall2017end}}.
However, the huge computation consumption and memory footprint caused by cost volume and 3D CNN of MVSNet may limit the potential of handling high-resolution image and the performance of 3d reconstruction.
Consequently, lots of efforts have been devoted to handling these issues, which can be divided into 2 categories: Recurrent-based methods \cite{yao2019recurrent, yan2020dense, wei2021aa} and Coarse-to-fine methods \cite{xiaodong2020cascade,chen2019point,yang2020cost,ma2021epp,zhu2021multi,cheng2020deep,yu2020fast,su2022uncertainty}.
The recurrent-based methods replace the 3D CNN with recurrent neural networks (RNN) to regularize the cost volume.
The coarse-to-fine methods separate the single-stage cost volume regularization process of MVSNet into multiple stages.

\subsection{Self-supervised Multi-view Stereo}
\label{sec:related:self_sup_mvs}

In aware of the expensive and time-consuming process for collecting ground truth depth maps in MVS tasks, a recent strand of work in unsupervised/self-supervised MVS methods strive to remove the reliance on ground truth and replace the depth regression loss with an image reconstruction loss built upon photometric consistency {\cite{khot2019learning}}.
In Unsup\_MVS {\cite{khot2019learning}}, the predicted depth map of MVSNet is used to reconstruct the reference image from source images via homography warping.
The self-supervised training follows the assumption that the reconstructed image from other views should be similar to the original image if the depth map is correct.
Although the self-supervision loss provides a promising alternative to supervised loss, it is not accurate enough and may be confused by many unexpected problems, such as occlusion ambiguity {\cite{dai2019mvs2}}, color constancy ambiguity {\cite{xu2021self}}, and textureless ambiguity {\cite{xu2021digging}}.

\subsection{Semi-supervised Learning and Multi-view Stereo}
\label{sec:related:semi_sup_mvs}

In recent years, immense progress has been witnessed in semi-supervised learning, especially in image classification.
Following the continuity assumption of semi-supervised learning {\cite{bachman2014learning,samuli2017temporal}}, \emph{consistency regularization} applies random data augmentation to semi-supervised learning by leveraging the idea that a classifier should output the same class distribution for an unlabeled example even after it has been augmented.
The basic consistency loss {\cite{sajjadi2016regularization}} in semi-supervised frameworks, such as $\Pi$-model {\cite{rasmus2015semi}}, Mean Teacher {\cite{tarvainen2017mean}}, Unsupvised Data Augmentation {\cite{xie2020unsupervised}} and MixMatch {\cite{berthelot2019mixmatch}} is the $l$-2 loss as follows:
\begin{equation}
\Omega (x ; \theta) = \| p_{\text{model}}(y | \text{perturb}(x) ; \theta) - p_{\text{model}}(y | x  ; \theta))\|_2^2
\label{eq_1}
\end{equation}
Note that $\text{perturb}(x)$ is a stochastic transformation, hence the two terms in Eq. \ref{eq_1} are not identical.
Consistency regularization enforces the unlabeled example $x$ to be classified the same as $\text{perturb}(x)$, a random augmentation of itself.
Whereas, different from the standard classification setting in semi-supervised learning, the Semi-MVS problem in this paper has to face huge variation of scenes in the MVS dataset, which may break the continuity assumption of labeled and unlabeled data distribution.
Consequently, further improvements are required in Semi-MVS problem.

The most recent work with a similar topic of semi-supervised setting in MVS, is SGT-MVSNet {\cite{kim2021just}}.
However, Sparse Ground-Truth based MVS (SGT-MVS) problem has a basic premise that the labeled pixels and unlabeled pixels are picked from the same images, which can be assumed to follow the same data distribution.
Different from SGT-MVS, the core problem of this work is how to propagate the supervision signal on labeled samples to unlabeled samples which may have a distribution gap.

\section{Method}
\label{sec:method}

\subsection{Problem Definition}
\label{sec:method:problem}

Given a pair of multi-view images with $N$ calibrated views, the reference image is denoted as $I_1$ and the $v$-th source view is denoted as ${\{I_v\}}_{v=2}^N$.
The intrinsic and extrinsic parameters on view $v$ are defined as ${\{K_v\}}_{v=1}^N$ and ${\{T_v\}}_{v=1}^N$ respectively.
The ground truth depth map on the reference view is noted as $D$.
A labeled sample is $S^l = \{ {\{ I_v^l, K_v^l, T_v^l \}}_{v=1}^N, D^l \}$ and an unlabeled sample is $S^u = \{ {\{ I_v^u, K_v^u, T_v^u \}}_{v=1}^N \}$.
Assume that $M$ samples are available in the whole MVS dataset, comprised of $\mu M$ labeled sample $S^l$ and $(1 - \mu) M$ unlabeled sample $S^u$.
Considering the difficulties in collecting dense depth ground truth, $\mu$ is set to a small ratio of $0.1$ in default, which creates a challenging task since only an extremely small ratio of ground truth is available.

\begin{figure*}[h] 
\centering 
\includegraphics[width=\textwidth]{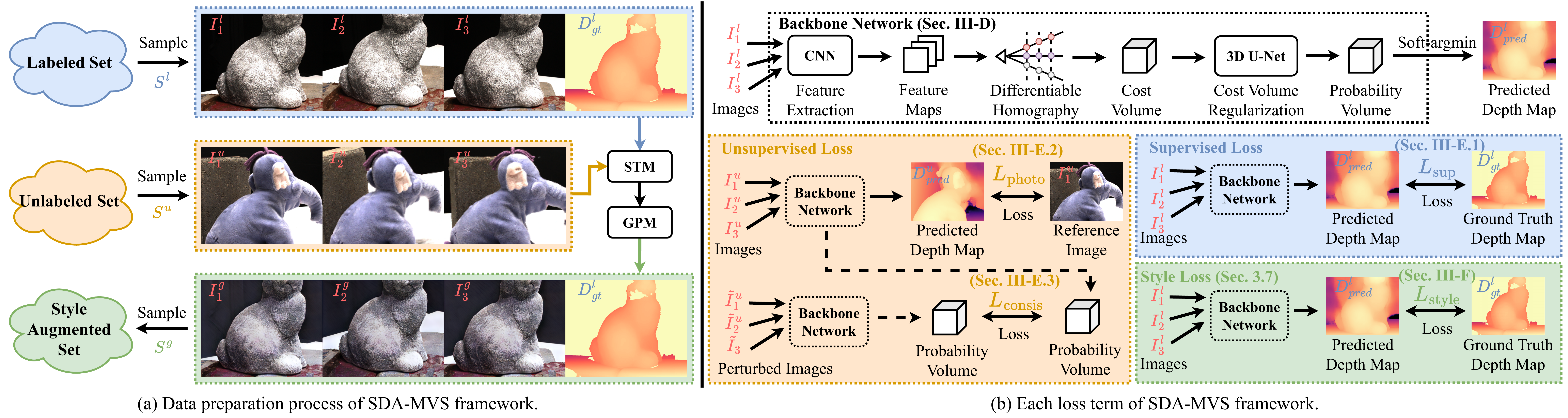} 
\caption{Overall framework of SDA-MVS framework.} 
\label{fig:framework} 
\end{figure*}

\subsection{Geometry-dropping Problem}


As discussed in Sec. \ref{sec:intro}, the distribution gap may break the semi-supervised assumption due to the huge variation among scenarios.
Taking inspiration from neural style transfer, we aim to transfer the visual style from unlabeled data to labeled data, trying to shrink this gap.
However, \emph{another problem of losing 3D geometric details} occurs when neural style transfer algorithms are directly applied to MVS images, as shown in Fig. \ref{fig:motivation} (b).
The results (2nd row) of STM embedded with standard neural style transfer algorithms lack geometric details on the zoomed region and the reconstructed 3D point cloud is much sparser compared with the original ones (1st row).
Putting the cart before the horse, the lost details in STM may reversely degrade the performance in MVS problem.
To handle this issue, we further utilize GPM to handle this problem as an image distortion problem.

\subsection{Overall Pipeline}
\label{sec:method:pipeline}

In Fig. \ref{fig:framework}, the overall framework of our proposed SDA-MVS is presented.
As shown in Fig. \ref{fig:framework}(a), labeled sample $S^l$ and unlabeled sample $S^u$ are randomly selected from the labeled and unlabeled dataset respectively in the data preparation process.
Then the labeled and unlabeled sample are fed to STM and GPM to generate style augmented sample $S^g$.
Afterwards, as shown in Fig. \ref{fig:framework}(b), the labeled sample $S^l$ is supervised under standard supervision loss (Eq. \ref{eq_2}).
The unlabeled sample $S^u$ is supervised under unsupervised loss (Eq. \ref{eq_5}) comprised of photometric consistency loss and consistency regularization loss.
The style augmented sample $S^g$ is enforced to satisfy the style consistency regularization framework (Sec. \ref{sec:method:style_consis}).

\subsection{Backbone}
\label{sec:method:backbone}

Arbitrary MVS network can be utilized as the backbone of the proposed semi-supervised framework, i.e. MVSNet {\cite{yao2018mvsnet}}, CasMVSNet {\cite{xiaodong2020cascade}}, and etc.
In default, the representative CasMVSNet {\cite{xiaodong2020cascade}} is used.
The MVS network requires $N$ multi-view images as input.
The feature map extracted by CNN with shared weights on each view is reprojected to the same reference view with differentiable homography warping.
The variance among the feature maps on different views is calculated to construct the cost volume, and a 3D U-Net is utilized to regularize the predicted probability volume $PV$.
The predicted depth map $D_{pred}$ is finally regressed with soft-argmin operation.

\subsection{Depth Consistency Regularization}
\label{sec:method:consis}

\subsubsection{Supervised Loss}
\label{sec:method:consis:sup}

The labeled sample is denoted as $S^l = \{ {\{ I_v^l, K_v^l, T_v^l \}}_{v=1}^N, D_{gt}^l \}$.
Following a standard supervised approach {\cite{yao2018mvsnet}}, the L2 loss between the predicted depth map $D_{pred}^l$ of the backbone network and the ground truth depth map $D^l$ on all valid pixels is minimized: 
\begin{equation}
    \small
    L_{sup} = \frac{\sum_{i=1}^{HW} \mathds{1} (D_{gt}^l(p_i) > 0) \| D_{pred}^l(p_i) - D_{gt}^l(p_i) \|_2^2} {\sum_{i=1}^{HW} \mathds{1} (D_{gt}^l(p_i) > 0)}
    \label{eq_2}
\end{equation}
where $i$ represents the index of available pixels in the $H \times W $ image, and $p_i$ is the pixel coordinate.
$\mathds{1} (D_{gt}^l(p_i) > 0)$ is the indicator function which represents whether valid depth ground truth exists in current pixel $p_i$.
Note that all invalid pixels in the provided ground truth depth map are set to 0.


\subsubsection{Photometric Consistency Loss}
\label{sec:method:consis:photo}

The unlabeled sample is denoted as $S^u = \{ {\{ I_v^u, K_v^u, T_v^u \}}_{v=1}^N \}$.
With the homography warping function, pixel $p_i^1$ in the reference image $I_1^u$ corresponds to pixel $\widehat{p}_i^{v}$ in the $v$-th source view image $I_v^u$. 
\begin{equation}
    \small
    D_v(\widehat{p}_i^{v}) \widehat{p}_i^{v} = K_v^u T_v^u (K_1^u T_1^u)^{-1} D_{pred}^u (p_i^1) p_i^1
    \label{eq_3}
\end{equation}
where $i (1 \leq i \leq HW)$ is the pixel index of $H \times W$ image.
$D_v$ represents the depth value on view $v$, and $D_{pred}^u$ is the predicted depth map from unlabeled sample $S^u$.
Since the $D_v(\widehat{p}_i^{v})$ is a scale term in homogeneous coordinates, we can normalize Eq. \ref{eq_3} to obtain the pixel coordinate $\widehat{p}_i^{v}$:
\begin{equation}
    \small
    \widehat{p}_i^{v} = \pi (D_v(\widehat{p}_i^{v}) \widehat{p}_i^{v}), \pi([x, y, z]^T) = [x/z, y/z, 1]^T
    \label{eq_4}
\end{equation}
With the correspondence relationship determined by Eq. \ref{eq_4}, the image on the reference view can be reconstructed via images on source view $v$:
$I_{v \rightarrow 1}^u(p_i^1) = I_v^u (\widehat{p}_i^{v})$.
Thus, the reconstructed image $I_{v \rightarrow 1}^u$ is enforced to be the same as original image $I_1^u$ following photometric consistency:
\begin{equation}
    \small
    L_{photo} = \sum_{j=2}^V \frac{\sum_{i=1}^{HW} \mathds{1}(1 \leq \widehat{p}_i^{v} \leq [H, W]) \| I_{v \rightarrow 1}^u (p_i) - I_1^u (p_i) \|_2^2}{\sum_{i=1}^{HW} \mathds{1}(1 \leq \widehat{p}_i^{v} \leq [H, W])}
    \label{eq_5}
\end{equation}
where $\mathds{1}(1 \leq \widehat{p}_i^{v} \leq [H, W])$ indicates whether the current pixel $p_i^1$ can find valid pixel $\widehat{p}_i^{v}$ in other source view.

\subsubsection{Consistency Regularization}
\label{sec:method:unsup:consis}

The general form of consistency regularization compute the divergence between the two predicted outputs of original sample and perturbed sample.
Denote that the perturbed version of unlabeled images $I_v^u$ is $\tilde{I}_v^u = \phi (I_v^u, \epsilon)$ by injecting a small noise $\epsilon$.
In MVS, the noise $\epsilon$ can be applied as hyperparameters controlling various data augmentation transformations like color jittering, gamma correction, image blurring and etc.
Similar as VAT {\cite{miyato2018virtual}}, we aim to minimize the KL divergence between the predicted distributions on an unlabeled sample ${\{I_v^u\}}_{v=1}^N$ and an augmented unlabeled sample ${\{\tilde{I}_v^u\}}_{v=1}^N$.


As a re-parametering trick, the soft-argmin operation {\cite{kendall2017end}} in the backbone network actually convert the discrete output of probability volume $PV$ into a continuous depth map by weighted summing it with all depth hypothesises.
Conversely, we can also treat the depth regression task in MVS as a classification task whose predicted classes are predefined depth space.
Assume that $K$ depth hypothesises are predefined in the MVS task, and the probability volume $PV$ with resolution of $H \times W \times K$ can be separated into $HW$ logits with $K$ categories.
In this way, we can simplify the dense depth regression problem into a per-pixel classification problem with $K$ predefined depth hypothesis(categories), and the probability volume is comprised of the predicted logits, which can be further used in the KL divergence based constraints as follows:
\begin{equation}
    \small
    L_{consis} = \frac{1}{HW} \sum_{i=1}^{HW} \mathbb{D}_{KL} (PV (p_i) || \widehat{PV} (p_i))
    \label{eq_6}
\end{equation}
where $\mathbb{D}_{KL}$ represents the KL divergence.
$i$ is the index of all $HW$ pixels in the image, and $p_i$ is the corresponding pixel coordinate.
$PV$ is the predicted probability volume of unlabeled sample ${\{I_v^u\}}_{v=1}^N$, and $\widehat{PV}$ is the predicted probability volume of augmented unlabeled sample ${\{\tilde{I}_v^u\}}_{v=1}^N$.
For multi-stage MVS network, only the initial stage is used to calculate the consistency regularization loss due to dynamic depth hypothesises in different stages.

\subsection{Style Consistency Regularization}
\label{sec:method:style_consis}

\subsubsection{Style Translation Module (STM)}

Based on aforementioned discussions, we aim to transfer the visual style of unlabeled image to labeled image, and shrink the distribution gap.
The basic assumption of neural style transfer {\cite{gatys2015neural}} is that the visual style is encoded by a set of Gram matrices ${\{G^{la}\}}_{la=1}^{La}$ where $G^{la} \in \mathbb{R}^{C_{la} \times C_{la}}$ is derived from the feature map $F^{la}$ of layer $la$ in a CNN by computing the correlation between activation channels:
$[ G^{la}(F^{la}) ]_{ij} = \sum_k F^{la}_{ik} F^{la}_{jk}$.
The Gram matrix captures semantic information which is irrelevant to position, and more likely to represent semantic visual styles {\cite{gatys2015neural}}.
For simplicity, we refer to a classic method called Whitening and coloring Transform (WCT {\cite{li2017universal}}) in STM.
WCT solve the style transfer problem with linear transforms on feature maps derived from Gram matrix, which can also be viewed as an eccentric covariance matrix.

Denote that the unlabeled sample image $I^u$ is viewed as style image and the labeled image $I^l$ is treated as content image.
Then the content feature map on layer $la$ of VGG {\cite{vgg}} is $F^{la}_c = F^{la}(I^l)$ and the style feature map is $F^{la}_s = F^{la}(I^u)$.
The general form of WCT is defined as follows:
\begin{equation}
    \small
    \widehat{F}{cs}^{la} = (E_s D_s^{\frac{1}{2}} E_s^T) (E_c D_c^{-\frac{1}{2}} E_c^T) F^{la}_c
    \label{eq_7}
\end{equation}
where $E_s D_s^{\frac{1}{2}} E_s^T$ is called coloring transform and $E_c D_c^{-\frac{1}{2}} E_c^T$ is called whitening transform.
$D_c$ and $E_c$ are respectively the diagonal matrix with eigenvalues and the corresponding orthogonal matrix with eigenvectors of covariance matrix $F_c^{la} {F_c^{la}}^T = E_c D_c E_c^T$.
In analogy, $D_s$ and $E_s$ represent eigenvalues and eigenvector of covariance matrix $F_s^{la} {F_s^{la}}^T = E_s D_s E_s^T$.
The intuition of whitening transform is to peel off the visual style defined by normalizing the content feature map $F_c^{la}$ while preserving the global content structure.
The intuition of coloring transform is the inverse process of whitening transform, and the visual styles of $F_s^{la}$ are appended to the whitened feature map whose visual style is peeled off in whitening transform.
By training an autoencoder on the images with the loss in Eq. \ref{eq_8}, the decoder is responsible for inverting transformed features back to the RGB space.
\begin{equation}
    \small
    I^u = \text{Dec}(F^{la}(I^u)), I^l = \text{Dec}(F^{la}(I^l))
    \label{eq_8}
\end{equation}
The decoder of autoencoder pretrained on the dataset can reconstruct the transformed feature map back into the style augmented image $I^g$:
\begin{equation}
    \small
    I^g = \text{Dec}(\widehat{F}{cs}^{la})
    \label{eq_9}
\end{equation}

\subsubsection{Geometry Preserving Module (GPM)}

As shown in Fig. \ref{fig:motivation} (b), directly applying neural style transfer algorithm may lose geometric details which are important for modeling 3D consistency among views in MVS.
The reason is that all operations of neural style transfer are processed on feature maps extracted by a VGG network, which is usually over 16 times smaller than the original image.
The detailed information modeling the local regions may be lost under such a small resolution, thus unexpected distortions may occur {\cite{li2017universal}}.
Consequently, to handle this issue, we utilize the spatial propagation network (SPN) {\cite{liu2017learning}} to filter the distortions in the image.
SPN is a generic framework that can be applied to many affinity-related tasks.
Here, we utilize SPN to model local pixel pairwise relationships, defined by the original image.
SPN has 2 branches: propagation network and guidance network.
In intuition, the weights of filters are learned through the CNN guidance network, which are further fed to propagation network to filter the distortions (Please refer to appendix for more details).
The training of the SPN requires original image $I = \{ I^u, I^l \}$ and reconstructed image with unexpected distortion $\text{Dec} (F^{la}(I))$. 
The original image is treated as a prior of local affinity and fed to the guidance network, while the distorted image $\text{Dec} (F^{la}(I))$ is fed to the propagation module in SPN.
The training loss for SPN is shown as follows:
\begin{equation}
\begin{aligned}
    \small
    L_{spn} = \frac{1}{N} \sum_{v=1}^N (\frac{1}{HW} \sum_{i=1}^{HW} \| I_v(p_i) - \widehat{I}_v(p_i) \|_2^2 + \\
    \frac{1}{|P_{sparse}|} \sum_{p_j \in P_{sparse}} \| I_1(p_j) - \widehat{I}_{v \rightarrow 1}(p_j) \|_2^2) 
    \label{eq_10}
\end{aligned}
\end{equation}
where the style transfered image is calculated by: $\widehat{I} = \text{SPN} (\text{Dec} (F^{la}(I)), I)$.
$P_{sparse}$ is the sparse point cloud extracted with COLMAP {\cite{schonberger2016pixelwise}} among the multi-view images.
Utilizing the sparse 3D points, corresponding on pixel $I_v$ is back-projected to pixel $p_j$ in reference view following homography warping function (Eq. \ref{eq_3}).
The sparse correspondence among views is enforced to retain the 3D consistency.
After training with Eq. \ref{eq_10}, the SPN is used to filter the style transferred image generated by Eq. \ref{eq_9}:
\begin{equation}
    \small
    \widehat{I}^g = \text{SPN} ( \text{Dec}(\widehat{F}{cs}^{la}), I^l)
    \label{eq_11}
\end{equation}

\subsubsection{Style Consistency Loss}

With the aforementioned modules, the visual style of unlabeled sample $S^u = \{ {\{ I_v^u, K_v^u, T_v^u \}}_{v=1}^N \}$ is transfered to labeled sample $S^l = \{ {\{ I_v^l, K_v^l, T_v^l \}}_{v=1}^N, D^l \}$, and the generated sample is noted as $S^g = \{ {\{ \widehat{I}_v^g, K_v^l, T_v^l \}}_{v=1}^N, D^l \}$.
The camera parameters and ground depth value of $S^g$ are shared with the original labeled sample $S^l$.
Following Eq. \ref{eq_7}, Eq. \ref{eq_9} and Eq. \ref{eq_11}, the generated image $\widehat{I}_v^g$ on each view $v$ is calculated by utilizing unlabeled image $I_1^u$ as style image and labeled image $I_v^l$ as content image.
Then the style augmented samples are fed to the backbone network and return the predicted depth map $D_{pred}^{g}$.
The style consistency loss requires the output depth map $D_{pred}^{g}$ of the style transferred samples $S^g$ to be the same as the ground truth $D^l$:
\begin{equation}
    \small
    L_{style} = \frac{\sum_{i=1}^{HW} \mathds{1} (D_{gt}^l(p_i) > 0) \| D_{pred}^g(p_i) - D_{gt}^l(p_i) \|_2^2} {\sum_{i=1}^{HW} \mathds{1} (D_{gt}^l(p_i) > 0)}
    \label{eq_12}
\end{equation}

\subsection{Overall Loss}
\label{sec_method_overall_loss}

As shown in Fig. \ref{fig:framework}, the overall loss is the sum of all aforementioned terms:
\begin{equation}
    \small
    L_{overall} = L_{sup} + L_{photo} + \lambda_1 * L_{consis} + \lambda_2 * L_{style}
    \label{eq_13}
\end{equation}
where $L_{sup}$(Eq. \ref{eq_2}) is the basic supervision loss on labeled sample $S^l$.
On unlabeled sample $S^u$, $L_{photo}$(Eq. \ref{eq_5}) is the basic photometric consistency loss in unsupervised MVS, and $L_{consis}$(Eq. \ref{eq_6}) is the consistency regularization loss.
$L_{style}$(Eq. \ref{eq_12}) is the style consistency calculated on style augmented sample $S^g$.
In default, $\lambda_1$ is set to $0.1$, and $\lambda_2$ is set to $1.0$.

\section{Experiment}
\label{sec:exp}

\subsection{Dataset and Evaluation Metric}
\label{sec:exp:dataset}

\textbf{DTU} (\emph{DTU}) \cite{jensen2014large}, \textbf{Tanks\&Temples} (\emph{T\&T}) \cite{knapitsch2017tanks}, \textbf{BlendedMVS} (\emph{BLD}) \cite{yao2020blendedmvs} and \textbf{GTA-SFM} (\emph{GTA}) \cite{wang2020flow} are utilized for evaluation in this paper.
For quantitative evaluation on \emph{DTU} benchmark, we calculate the \emph{accuracy} and the \emph{completeness} predefined by official protocol.
The \emph{overall} score takes the average of mean accuracy and mean completeness as the reconstruction quality.
For quantitative evaluation on \emph{T\&T} benchmark, we evaluate the intermeidate and advanced set  according to their online benchmark.
For the \emph{BLD} and \emph{GTA} datasets, we implement the quantitative evaluation ourselves which also supports GPU parallel computation.
The ground truth point clouds are sampled from the provided meshes, and the evaluation protocol follows the metric of \emph{precision}, \emph{recall}, and \emph{f-score} defined in \emph{T\&T}.

\begin{table*}[t]
\centering
\caption{Ablation study of the proposed method on DTU, BlendedMVS and GTASFM datasets. $\uparrow$ means the higher the better, and $\downarrow$ means the lower the better.}
\resizebox{0.8\hsize}{!}{
\begin{tabular}{c|cccc|ccc|ccc|ccc}
\hline
\multirow{2}{*}{3D Anno.}       & \multicolumn{4}{c|}{Loss}        & \multicolumn{3}{c|}{\emph{DTU} Dataset} & \multicolumn{3}{c|}{\emph{BLD} Dataset} & \multicolumn{3}{c}{\emph{GTA} Dataset} \\
    & $L_{photo}$  & $L_{consis}$   & $L_{style}$   & $L_{sup}$    & Acc. $\downarrow$      & Comp.  $\downarrow$   & \textbf{Overall} $\downarrow$   & Prec.  $\uparrow$    & Reca. $\uparrow$    & \textbf{F-score} $\uparrow$   & Prec.  $\uparrow$   & Reca.  $\uparrow$   & \textbf{F-score}  $\uparrow$  \\ \hline
0\%                          & \checkmark &  &  &  & 0.3748     & 0.3598    & 0.3673     & 0.3528     & 0.1575    & 0.1796     & 0.4055    & 0.4441    & 0.4170      \\ \hline
\multirow{2}{*}{10\%-S} & \checkmark & \checkmark &  &  & 0.3572     & 0.3603    & 0.3588     & 0.4064     & 0.4135    & 0.3632     & 0.4264    & 0.4392    & 0.4255     \\
                             & \checkmark & \checkmark & \checkmark &  & 0.3447     & 0.3560    & 0.3504     & 0.4217     & 0.4328    & 0.3806     & 0.4523    & 0.5286    & 0.4807     \\ \hline
\multirow{2}{*}{10\%-V}  & \checkmark & \checkmark &     &  & 0.3498     & 0.3480    & 0.3489     & 0.4269     & 0.4475    & 0.3857     & 0.4870    & 0.6089    & 0.5351     \\
                             & \checkmark & \checkmark & \checkmark &  & 0.3305     & \textbf{0.3369}    & \textbf{0.3337}     & \textbf{0.4446}     & \textbf{0.4614}    & \textbf{0.4033}     & \textbf{0.5354}    & \textbf{0.6261}    & \textbf{0.5711}     \\ \hline
100\%                        &   &       &     & \checkmark & \textbf{0.325}      & 0.385     & 0.355      & 0.3644     & 0.4491    & 0.3760     & 0.5239    & 0.4794    & 0.4796     \\ \hline
\end{tabular}
}
\label{tab:abla_different_datasets}
\end{table*}

\begin{figure*}[t]
\centering
\includegraphics[width=0.72\textwidth]{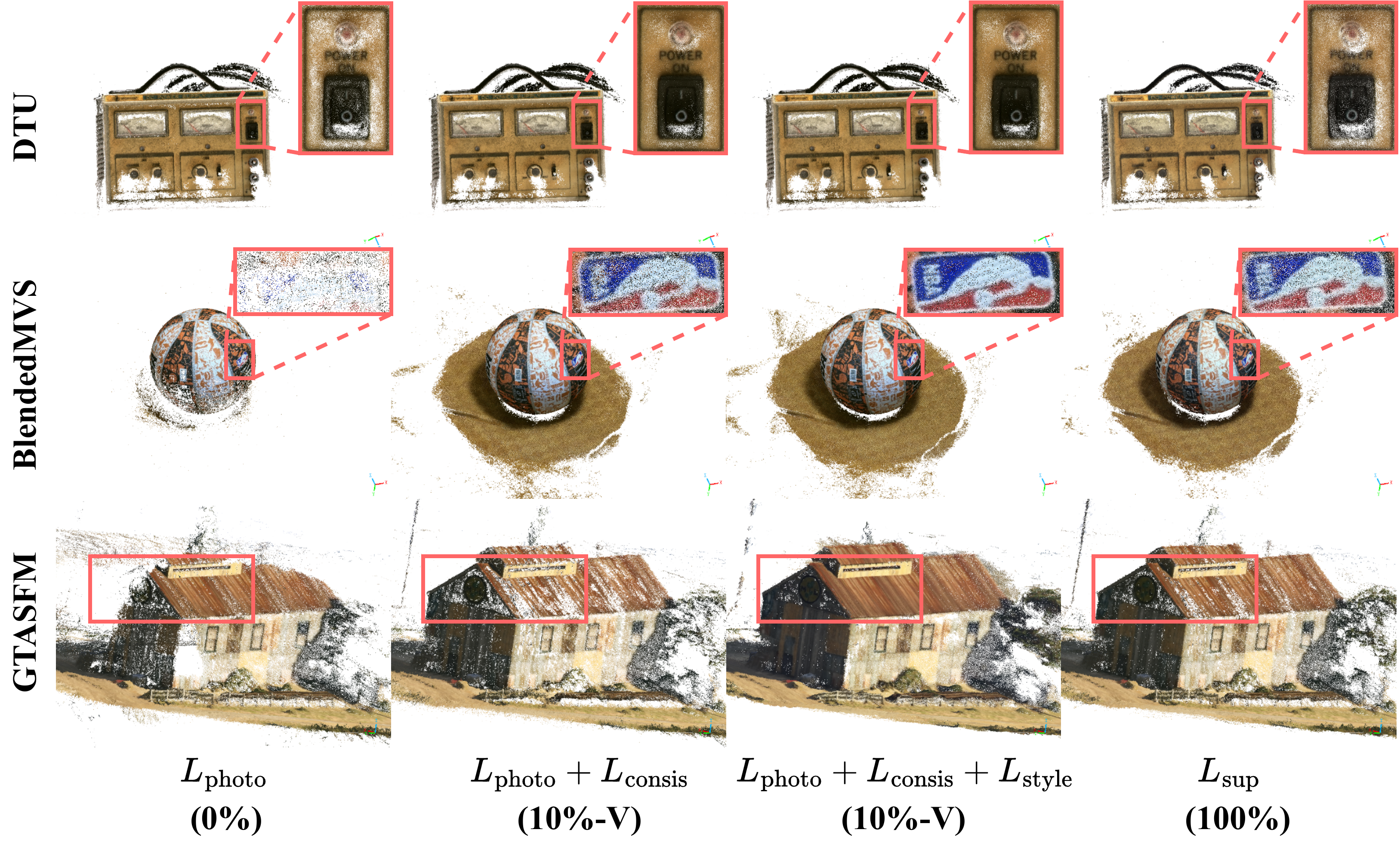} 
\caption{Qualitative ablation study of our proposed method under the setting of 10\%-V.}
\label{fig:qual_ablation_10v}
\end{figure*}

\begin{figure*}[t]
\centering
\includegraphics[width=0.72\textwidth]{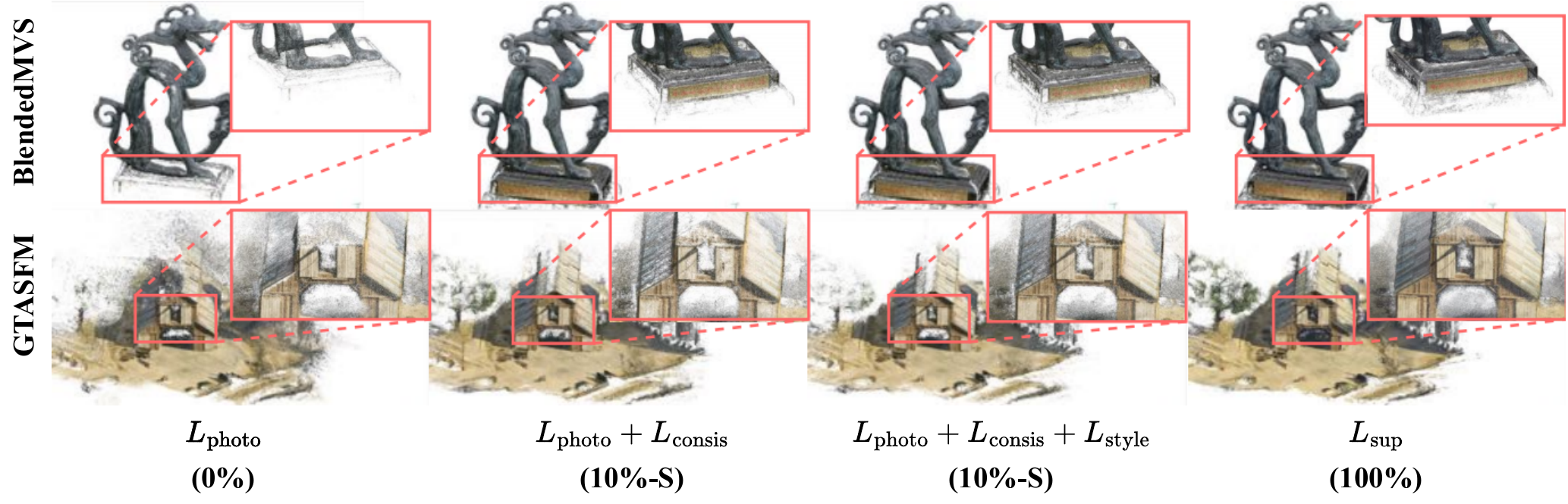} 
\caption{Qualitative ablation study of our proposed method under the setting of 10\%-S.}
\label{fig:qual_ablation_10s}
\end{figure*}

\begin{figure}[t]
\centering
\includegraphics[width=\hsize]{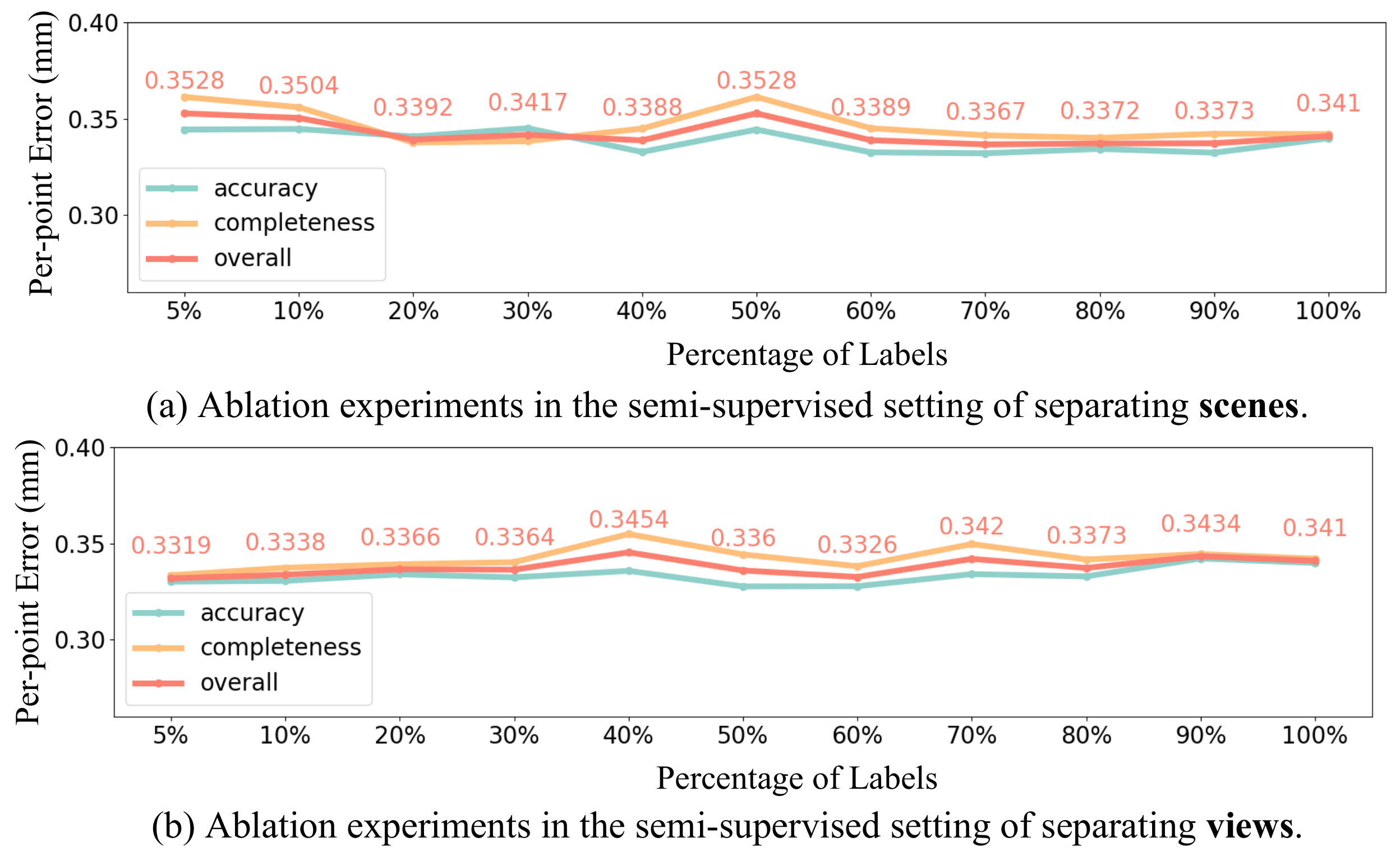} 
\caption{The performance of our proposed semi-supervised MVS method under different percentage of labeled data on DTU benchmark in the setting of separating views.}
\label{fig:abla_percentage_10s}
\label{fig:abla_percentage_10v}
\end{figure}

\begin{table}[t]
    \centering
    \caption{The performance of our proposed method under 5 different random splits.}
    \resizebox{!}{0.14\hsize}{
    \begin{tabular}{c|ccc|ccc}
    \hline
    Settings & \multicolumn{3}{c|}{10\%-S} & \multicolumn{3}{c}{10\%-V} \\ \hline
    ID       & Acc.    & Comp.   & Overall & Acc.    & Comp.  & Overall \\ \hline
    1                        & 0.3305  & 0.3369  & 0.3337  & 0.3447  & 0.3560 & 0.3504  \\
    2                        & 0.3370  & 0.3379  & 0.3375  & 0.3519  & 0.3416 & 0.3468  \\
    3                        & 0.3280  & 0.3371  & 0.3326  & 0.3594  & 0.3486 & 0.3540  \\
    4                        & 0.3279  & 0.3357  & 0.3318  & 0.3533  & 0.3576 & 0.3555  \\
    5                        & 0.3323  & 0.3399  & 0.3361  & 0.3567  & 0.3589 & 0.3578  \\ \hline
    \end{tabular}
    }
   
    \label{tab:random_sample}
\end{table}




\subsection{Implementation Detail}
\label{sec:exp:impl}

\noindent\textbf{Training and testing:}
In default, we utilize CasMVSNet \cite{xiaodong2020cascade} as the backbone network.
The split of train, valid and test sets in each dataset follows the official configuration in \emph{DTU}, \emph{BLD} and \emph{GTA}.
On each dataset, the model is trained on the training set first and tested on the test set for ablation study and evaluation.
8 NVIDIA V100 GPUs are used for training.
The batch size on each GPU is set to 1 and the training procedure requires 16 epochs.
The number of views is set to 3 following the default setting of CasMVSNet \cite{xiaodong2020cascade}.
The hyper-parameter setting of backbone networks follows the official CasMVSNet as well.
In default, Whitening and Coloring Transform (WCT \cite{li2017universal}) is selected as the backbone of STM module.
The backbone of GPM is a spatial propagation network \cite{liu2017learning}, a generic framework that can be applied to many affinity-related tasks, including but not limited to image matting, segmentation and etc.
STM and GPM are further trained on each dataset following Eq. \ref{eq_8} and Eq. \ref{eq_10} respectively.
For evaluation, the model is tested on the evaluation or test set of \emph{DTU}, \emph{T\&T}, \emph{BLD}, and \emph{GTA}.

\noindent\textbf{Semi-supervised settings:}
2 different semi-supervised settings are adopted in experiments: (1) labeled and unlabeled data separated by \textbf{scenes}; (2) labeled and unlabeled data separated by \textbf{views}.
From the visualization results in Figure \ref{fig:motivation}(a), it is verified that apparent gap of distribution exists among different scenes.
The former setting based on scenes aims at exploring the performance of our proposed method under this challenging case.
For the latter one, we aim to simulate the common case that labeled and unlabeled data are separated by the multi-view image pairs on different views.


\subsection{Ablation Study}
\label{sec:exp:abla}

\subsubsection{Loss terms of SDA-MVS}
\label{sec:exp:abla:loss}

To validate the effectiveness of each loss term in the proposed method, we conduct ablation study about the loss terms in this section.
Experiments on \emph{DTU}, \emph{BLD} and \emph{GTA} are conducted on the model trained with the training set of each dataset and tested on the corresponding test set.
The backbone model is CasMVSNet \cite{xiaodong2020cascade}.
The quantitative ablation results are shown in Tab. \ref{tab:abla_different_datasets}.
Different loss terms are used respectively for experiments, including $L_{\text{photo}}$, $L_{\text{consis}}$, $L_{\text{style}}$, and $L_{\text{sup}}$.
In the table, the first column means the utilized setting of 3D annotations.
0\% and 100\% respectively represent the unsupervised baseline and the fully-supervised baseline.
Following the 2 semi-supervised settings defined in Sec. \ref{sec:exp:impl}, we randomly select 10\% samples with 3D ground truth from the whole dataset and the remaining 90\% are unlabeled samples.
10\%-S means 10\% labeled samples are selected based on scenes, 10\%-V means 10\% labeled sample are selected based on views.
From the table, we can find that our semi-supervised method achieves competitive performance on the \textbf{Overall} score of \emph{DTU}, the \textbf{F-score} of \emph{BLD} and \emph{GTA}, compared with the supervised baselines.
As shown, our method under the setting of 10\%-V achieves \textbf{Overall} metric of 0.3337, which is better than the score of 0.3673 in unsupervised baseline and the score of 0.355 in fully-supervised baseline.
Similar results are also provided under the setting of 10\%-S, where our method shows competitive performance with the \textbf{Overall} metric of 0.3504 compared to the supervised baseline and unsupervised baseline.
Furthermore, qualitative results of the ablation experiments under the settings of 10\%-V and 10\%-S are respectively shown in Fig. \ref{fig:qual_ablation_10v} and Fig. \ref{fig:qual_ablation_10s}.

\begin{table}[t]
\centering
\caption{Quantitative results on DTU evaluation set (Lower is better). \underline{CasMVSNet} is selected as the backbone and the fully-supervised baseline in our SDA-MVS framework. The photometric consistency loss of \underline{Unsup\_MVS} is used to train the unsupervised baseline.}
\resizebox{!}{0.6\hsize}{
\begin{tabular}{c|l|lll}
\hline
Anno.                                  & Method        & Acc.           & Comp.          & Over.        \\ \hline 
\multirow{4}{*}{Trad.}   & Furu  {\cite{furukawa2009accurate}}        & 0.613          & 0.941          & 0.777          \\
                         & Tola  {\cite{tola2012efficient}}        & 0.342          & 1.190           & 0.766          \\
                         & Camp  {\cite{campbell2008using}}        & 0.835          & 0.554          & 0.694          \\
                         & Gipuma  {\cite{galliani2015massively}}      & \textbf{0.283} & 0.873          & 0.578          \\ 
                         & Colmap  {\cite{schonberger2016pixelwise}} & 0.400 & 0.644 & 0.532 \\\hline
\multirow{15}{*}{100\%}   & Surfacenet  {\cite{ji2017surfacenet}}   & 0.450           & 1.040           & 0.745          \\
                         & MVSNet  {\cite{yao2018mvsnet}}      & 0.396          & 0.527          & 0.462          \\
                         & CIDER  {\cite{xu2020learning}}        & 0.417          & 0.437          & 0.427          \\
                         & P-MVSNet  {\cite{luo2019p}}    & 0.406          & 0.434          & 0.420           \\
                         & R-MVSNet  {\cite{yao2019recurrent}}    & 0.383          & 0.452          & 0.417          \\
                         & MVSNet++ {\cite{chen2020mvsnet++}}   & 0.345 & 0.407 & 0.376 \\
                         & Point-MVSNet  {\cite{chen2019point}} & 0.342          & 0.411          & 0.376          \\
                         & Fast-MVSNet  {\cite{yu2020fast}}   & 0.336          & 0.403          & 0.370           \\
                         & \underline{CasMVSNet}  {\cite{xiaodong2020cascade}}     & 0.325          & 0.385          & 0.355          \\
                         & UCS-Net  {\cite{cheng2020deep}}     & 0.330           & 0.372          & 0.351          \\
                         & CVP-MVSNet  {\cite{yang2020cost}}   & 0.296          & 0.406          & 0.351          \\
                         & PatchMatchNet  {\cite{wang2021patchmatchnet}} & 0.427          & 0.277          & 0.352          \\
                         & AA-RMVSNet  {\cite{wei2021aa}}   & 0.376          & 0.339          & 0.357          \\
                         & EPP-MVSNet  {\cite{ma2021epp}}   & 0.413          & 0.296          & 0.355          \\
                         & UGNet {\cite{su2022uncertainty}} & 0.334 & 0.330 & 0.332 \\
                         & MVSTR  {\cite{zhu2021multi}}        & 0.356          & \textbf{0.295} & \textbf{0.326} \\
                         & PVSNet  {\cite{xu2020pvsnet}}       & 0.337          & 0.315          & \textbf{0.326} \\ \hline
\multirow{7}{*}{0\%}  & \underline{Unsup\_MVS}  {\cite{khot2019learning}}   & 0.881          & 1.073          & 0.977          \\
                         & MVS$^2$  {\cite{dai2019mvs2}}        & 0.760           & 0.515          & 0.637          \\
                         & M$^3$VSNet  {\cite{huang2021m3vsnet}}      & 0.636          & 0.531          & 0.583          \\
                         & Meta\_MVS  {\cite{mallick2020learning}}    & 0.594          & 0.779          & 0.687          \\
                         & JDACS   {\cite{xu2021self}}     & 0.571          & 0.515          & 0.543          \\
                         & JDACS-MS   {\cite{xu2021self}}    & 0.398          & 0.318          & 0.358          \\
                         & U-MVS  {\cite{xu2021digging}}       & 0.354          & 0.354         & 0.354         \\ \hline
Sparse                  & SGT-MVS  {\cite{kim2021just}}   & 0.421         & 0.349         & 0.385         \\ \hline
10\%-S                & SDA-MVS (ours)    & 0.345         & 0.356          & 0.350         \\
10\%-V                & SDA-MVS (ours)    & 0.331         & 0.337         & 0.334         \\ \hline
\end{tabular}}
\label{tab:quan_dtu}
\end{table}

\begin{table*}[t]
\centering
\caption{Quantitative results of different methods on Tanks and Temples benchmark (higher is better). \underline{CasMVSNet} represents the fully-supervised baseline.}
\resizebox{0.85\hsize}{!}{
\begin{tabular}{c|ccccccccc|ccccccc}
\hline
 \multirow{2}{*}{\diagbox{Method}{F-Score}}  & \multicolumn{9}{c|}{\emph{T\&T} Intermediate}    & \multicolumn{7}{c}{\emph{T\&T} Advanced} \\  
        & Fam.  & Franc.    & Horse & Light.    & M60   & Pan.  & Play. & Train & \textbf{Mean} & Audi. & Ballr.    & Courtr.   & Museum    & Palace    & Temple    & \textbf{Mean} \\ \hline 
OpenMVG + PMVS {\cite{furukawa2009accurate}}    & 41.03 &	17.70 &	12.83 &	36.68 &	35.93 &	33.20 &	31.78 &	28.10 & 29.66 & 4.54 &	12.09 &	21.00 &	29.17 &	6.76 & 12.72 & 14.38 \\
OpenMVG + SMVS {\cite{langguth2016shading}}    & 31.93 &	19.92 &	15.02 &	39.38 &	36.51 &	41.61 &	35.89 &	25.12 & 30.67 & 6.96 &	11.58 &	19.82 &	21.89 &	8.90 & 12.27 & 13.57 \\
OpenMVG + MVE {\cite{fuhrmann2014mve}}    & 49.91 &	28.19 &	20.75 &	43.35 &	44.51 &	44.76 &	36.58 &	35.95 & 38.00 & 14.70 &	26.36 &	32.48 &	37.57 &	3.65 & 22.84 & 22.93 \\
COLMAP {\cite{schonberger2016pixelwise}} & 42.14 & 50.41     & 22.25 & 25.63     & 56.43 & 44.83 & 46.97 & 48.53 & 42.04         & 27.24 & 16.02     & 25.23     & 34.7      & 41.51     & 18.05     & 27.94            \\ 
\hline
MVSNet {\cite{yao2018mvsnet}}       & 55.99          & 28.55          & 25.07          & 50.79          & 53.96          & 50.86          & 47.9           & 34.69          & 43.48          & \textbackslash & \textbackslash & \textbackslash & \textbackslash & \textbackslash & \textbackslash & \textbackslash \\
R-MVSNet {\cite{yao2019recurrent}}     & 69.96          & 46.65          & 32.59          & 42.95          & 51.88          & 48.8           & 52             & 42.38          & 48.4           & 12.55            & 29.09            & 25.06            & 38.68            & 19.14            & 24.96            & 24.91            \\
CIDER {\cite{xu2020learning}}        & 56.79          & 32.39          & 29.89          & 54.67          & 53.46          & 53.51          & 50.48          & 42.85          & 46.76          & 12.77            & 24.94            & 25.01            & 33.64            & 19.18            & 23.15            & 23.12            \\
Fast-MVSNet {\cite{yu2020fast}}  & 65.18          & 39.59          & 34.98          & 47.81          & 49.16          & 46.2           & 53.27          & 42.91          & 47.39          & \textbackslash & \textbackslash & \textbackslash & \textbackslash & \textbackslash & \textbackslash & \textbackslash \\
\underline{CasMVSNet} {\cite{xiaodong2020cascade}}    & 76.37          & 58.45          & 46.26          & 55.81          & 56.11          & 54.06          & 58.18          & 49.51          & 56.84          & 19.81            & 38.46            & 29.1             & 43.87            & 27.36            & 28.11            & 31.12            \\
UCS-Net {\cite{cheng2020deep}}      & 76.09          & 53.16          & 43.03          & 54             & 55.6           & 51.49          & 57.38          & 47.89          & 54.83          & \textbackslash & \textbackslash & \textbackslash & \textbackslash & \textbackslash & \textbackslash & \textbackslash \\
CVP-MVSNet {\cite{yang2020cost}}   & 76.5           & 47.74          & 36.34          & 55.12          & 57.28          & 54.28          & \textbf{57.43} & 47.54          & 54.03          & \textbackslash & \textbackslash & \textbackslash & \textbackslash & \textbackslash & \textbackslash & \textbackslash \\
PVANet {\cite{yi2020pyramid}}       & 69.36          & 46.8           & 46.01          & 55.74          & 57.23          & \textbf{54.75} & 56.7           & 49.06          & 54.46          & \textbackslash & \textbackslash & \textbackslash & \textbackslash & \textbackslash & \textbackslash & \textbackslash \\
PatchmatchNet {\cite{wang2021patchmatchnet}} & 66.99          & 52.64          & 43.24          & 54.87          & 52.87          & 49.54          & 54.21          & 50.81 & 53.15          & \textbf{23.69}   & 37.73            & 30.04            & 41.8             & 28.31            & 32.29            & 32.31            \\
MVSTR {\cite{zhu2021multi}}        & 76.92          & \textbf{59.82} & 50.16          & \textbf{56.73} & 56.53          & 51.22          & 56.58          & 47.48          & 56.93          & 22.83            & \textbf{39.04}   & \textbf{33.87}   & \textbf{45.46}   & 27.95            & 27.97            & \textbf{32.85}   \\ \hline
MVS$^2$ {\cite{dai2019mvs2}} & 47.74 & 21.55 & 19.50 & 44.54 & 44.86 & 46.32 & 43.38 & 29.72 & 37.21 & \textbackslash & \textbackslash & \textbackslash & \textbackslash & \textbackslash & \textbackslash & \textbackslash \\
M$^3$VSNet {\cite{huang2021m3vsnet}} & 47.74 & 24.38 & 18.74 & 44.42 & 43.45 & 44.95 & 47.39 & 30.31 & 37.67 & \textbackslash & \textbackslash & \textbackslash & \textbackslash & \textbackslash & \textbackslash & \textbackslash \\
JDACS {\cite{xu2021self}} & 66.62 & 38.25 & 36.11 & 46.12 & 46.66 & 45.25 & 47.69 & 37.16 & 45.48 & \textbackslash & \textbackslash & \textbackslash & \textbackslash & \textbackslash & \textbackslash & \textbackslash \\
U-MVS {\cite{xu2021digging}} & 76.49 & 60.04 & 49.20 & 55.52 & 55.33 & 51.22 & 56.77 & 52.63 & 57.15 & 22.79 & 35.39 & 28.90 & 36.70 & 28.77 & 33.25 & 30.97 \\
\hline
SDA-MVS(10\% Scenes)     & \textbf{77.19} & 59.80          & \textbf{52.87} & 54.63          & 57.11 & 52.10           & 55.86          & \textbf{51.63}          & \textbf{57.65} & 21.70            &  36.45           & 30.15            &  37.27           &  \textbf{29.01}  & \textbf{34.54}   & 31.52 \\
SDA-MVS(10\% Views)         & 77.09 & 55.55          & 52.59 & 55.66          & \textbf{58.17} & 51.7           & 55.58          & 50.64          & 57.12 & 22.62            & 37.73            & 29.51            & 37.34            & 28.95   & 34.33   & 31.74          \\ \hline
\end{tabular}
}
\label{tab:quan_tt}
\end{table*}

\subsubsection{Percentage of Labels}
\label{sec:exp:abla:percentage}

The percentage of labeled samples in the semi-supervised setting is an important factor to demonstrate the effectiveness of the proposed method.
To explore the effectiveness of our proposed method under different percentage of labeled samples, we conduct experiments under the labeled percentage ranging from 5\% to 100\%.
As shown in Fig. \ref{fig:abla_percentage_10s}, we report the per-point errors of reconstructed dense 3D point cloud under the percentage of 5\%, 10\%, 20\%, 30\%, 40\%, 50\%, 60\%, 70\%, 80\%, 90\%  and 100\%.
Each percentage setting is randomly selected from all available samples for a fair comparison.
However, the random selection process may make the performance fluctuate to some extent, which is further discussed detailly in Sec. \ref{sec:exp:abla:random_sample}.
Here, we can only summarize the tendency of increasing the percentage of labeled samples from the experiment results.
Note that the 100\% setting here is the combination of fully-supervised training and our proposed framework.
Compared with the fully-supervised baseline whose \textbf{Overall} score is about 0.355, each percentage setting of our proposed framework in Fig. \ref{fig:abla_percentage_10s} and Fig. \ref{fig:abla_percentage_10v} achieves superior performance which is smaller than 0.355.
It demonstrates the stable improvement of our proposed method compared to the basic fully-supervised baseline as well as the unsupervised one.

\subsubsection{Random Sampling of Labeled and Unlabeled data}
\label{sec:exp:abla:random_sample}

Due to the random sampling process of labeled and unlabeled data, the unexpected factors in random separation might be a counfounder to the exact performance of our semi-supervised MVS method.
To prove the exact performance under different random selected samples, we conduct random sampling repeatedly for 5 times and provide the benchmarking results on \emph{DTU} in Tab. \ref{tab:random_sample}.
As the results show, the \emph{Overall} metric of 10\%-V fluctuates around 0.33 and the \emph{Overall} metric of 10\%-S fluctuates around 0.352.
These results prove that our semi-supervised method can achieve competitive and even better performance compared with unsupervised and fully-supervised baseline.

\subsection{Benchmark Result}

\subsubsection{Results on DTU}

We compare the proposed semi-supervised framework with other state-of-the-art methods quantitatively in Table \ref{tab:quan_dtu}
The 1-st column of the table means the annotation strategy of the method: 1) \emph{Trad.} means traditional MVS methods without annotation; 2) \emph{100\%} means fully-supervised MVS methods; 3) \emph{0\%} means unsupervised MVS methods; 4) \emph{Sparse} means sparse point annotations according to \cite{kim2021just}; 5) \emph{10\%-S} and \emph{10\%-V} means the aforementioned semi-supervised settings (10\% scenes / views) in Sec. \ref{sec:exp:impl}.
In the table, we compare our method with state-of-the-art traditional MVS methods (i.e. Furu  \cite{furukawa2009accurate}, Gipuma  \cite{galliani2015massively}, and Colmap  \cite{schonberger2016pixelwise}), learning-based MVS methods(i.e. MVSNet  \cite{yao2018mvsnet}, R-MVSNet  \cite{yao2019recurrent}, Point-MVSNet  \cite{chen2019point}, CasMVSNet \cite{xiaodong2020cascade}, UCS-Net  \cite{cheng2020deep}, and UGNet \cite{su2022uncertainty}), and unsupervised MVS methods (i.e. Unsup\_MVS  \cite{khot2019learning}, M$^3$VSNet  \cite{huang2021m3vsnet}, JDACS   \cite{xu2021self}, and U-MVS  \cite{xu2021digging}).
Furthermore, we also compare the recent semi-supervised MVS methods, SGT-MVS \cite{kim2021just}.
As shown in the table, with limited 10\% dense ground truth in the training set, our proposed method performs competitively compared with supervised MVS methods, achieving an overall score of 0.334.
Furthermore, compared with the reported official supervised performance of the backbone network, CasMVSNet   \cite{xiaodong2020cascade}, our proposed method achieve better performance with much less dense 3D annotations.
In addition, the proposed SDA-MVS outperforms previous state-of-the-art traditional MVS methods and unsupervised MVS methods are presented in the table.

\begin{table}[t]
\centering
\caption{Experimental results under the setting of unsupervised domain adaptation.}
\resizebox{!}{0.10\hsize}{
\begin{tabular}{c|c|ccc|ccc}
\hline
\multirow{2}{*}{}             & \multirow{2}{*}{Methods} & \multicolumn{3}{c|}{\emph{BLD} $\rightarrow$ \emph{DTU}} & \multicolumn{3}{c}{\emph{GTA} $\rightarrow$ \emph{DTU}} \\
                              &                          & Acc.         & Comp.        & Overall      & Acc.         & Comp.       & Overall      \\ \hline
Source Only                   & CasMVSNet \cite{xiaodong2020cascade}           & 0.3609       & 0.4024       & 0.3817       & 0.4131       & 0.8854      & 0.6493       \\ \hline
\multirow{3}{*}{SSDA(10\%-S)} & $\Pi$-model {\cite{rasmus2015semi}}     & 0.3636       & 0.3815       & 0.3726       & 0.3767       & 0.549       & 0.4629       \\
                              & Zhang et al. \cite{zhang2020label}            & 0.3619       & 0.3634       & 0.3627       & 0.3766       & 0.405       & 0.3908       \\
                              & SDA-MVS                  & 0.3681       & 0.3488       & 0.3585       & 0.3515       & 0.3801      & 0.3658       \\ \hline
\end{tabular}
}
\label{tab:ssda}
\end{table}

\subsubsection{Results on Tanks\&Temples}

Leveraging the model trained by the training set of \emph{DTU} without finetuning, we test the performance of our proposed method on \emph{T\&T}.
In Tab. \ref{tab:quan_tt}, the results on the \emph{Intermediate} and \emph{Advanced} set of \emph{T\&T} are reported.
The reported F-score on both the intermediate and advanced partitions are used in the table.
We compare our proposed method with traditional methods (i.e. PMVS \cite{furukawa2009accurate}, SMVS \cite{langguth2016shading}, MVE \cite{fuhrmann2014mve}, COLMAP \cite{schonberger2016pixelwise}), fully-supervised MVS methods (i.e. MVSNet \cite{yao2018mvsnet}, R-MVSNet \cite{yao2019recurrent}, CasMVSNet \cite{xiaodong2020cascade}, UCS-Net \cite{cheng2020deep}, PatchmatchNet \cite{wang2021patchmatchnet}, MVSTR \cite{zhu2021multi}), and self-supervised MVS methods (i.e. MVS$^2$ \cite{dai2019mvs2}, M$^3$VSNet \cite{huang2021m3vsnet}, JDACS \cite{xu2021self}, U-MVS \cite{xu2021digging}).
Note that our SDA-MVS framework only applies the simplest self-supervision loss based on Unsup\_MVS \cite{khot2019learning} rather than the state-of-the-art self-supervised methods like U-MVS \cite{xu2021digging} to maintain a simple and compact framework.
The results demonstrate that our proposed method can outperform its supervised and unsupervised baseline significantly, and providing competitive performance compared with other state-of-the-art methods.

\subsubsection{Unsupervised Domain Adaptation on MVS}

Recent work \cite{zhang2020label} reveals that semi-supervised learning (SSL) can be simply extended to unsupervised domain adaptation (UDA) and achieve great performance.
We further evaluate the proposed SDA-MVS by extending it to the unsupervised domain adaptation task and present the experimental results in Tab. \ref{tab:ssda}.
The CasMVSNet  \cite{xiaodong2020cascade} is selected as the backbone network.
We select \emph{BLD}/\emph{GTA} as the source domain, and \emph{DTU} as the target domain, denoted as \emph{BLD}/\emph{GTA} $\rightarrow$ \emph{DTU}.
As shown in the table, our method can also achieve great performance when generalizing to the unsupervised domain adaptation MVS task.
It reveals the extensive potential of our proposed SDA-MVS from SSL to UDA.

\section{Conclusion}
\label{sec:conclu}

In this paper, we explore the semi-supervised MVS problem that assumes only part of the MVS dataset has dense depth annotations.
It has an intractable risk of breaking the basic assumption in classic semi-supervised learning techniques, that labeled data and unlabeled data share same label space and data distribution, denoted as \emph{semi-supervised distribution-gap ambiguity}.
To handle these issues, we propose a novel semi-supervised distribution-augmented MVS framework, called SDA-MVS.
For the case that the assumption works in the MVS data, consistency regularization based on the KL divergence between the predicted probability volumes with and without random data augmentation is enforced to train the model.
For the case that the assumption fails in the MVS data because of distribution mismatch, style consistency regularization enforces the invariance between the style augmented sample and original labeled sample.
The style augmented sample is generated by transferring visual styles from unlabeled data to labeled data, inherently shrinking the distribution gap.
Experimental results show that our proposed SDA-MVS can handle the semi-supervised MVS effectively and be extended to an UDA MVS with great performance.

\begin{acks}
This work was supported by the National Natural Science Foundation of China (No.61976095) and the Natural Science Foundation of Guangdong Province, China (No.2022A1515010114).
This work was also supported by Alibaba Group through Alibaba Research Intern Program.
\end{acks}

\section{Appendix}

\appendix

\section{Implementation Details}

Here we provide a more detailed version of the implementation details.

\subsection{Datasets}
\label{sec:exp:datasets}

\noindent\textbf{DTU} (\emph{DTU}) {\cite{jensen2014large}} is an indoor multi-view stereo dataset with 128 different scenes along with 7 different lighting conditions. Each scene is attached with a ground truth point cloud and multi-view images captured from 49 or 64 fixed viewpoints. 
We follow the same configuration of train, valid, and test set provided by Yao et al. {\cite{yao2018mvsnet}} for a fair comparison.

\noindent\textbf{Tanks\&Temples} (\emph{T\&T}) {\cite{knapitsch2017tanks}} is a large-scale outdoor MVS dataset that consists of various challenging scenarios. 
Following the official online benchmark, the intermediate and advanced partition of Tanks\&Temples benchmark is used for evaluation.

\noindent\textbf{BlendedMVS} (\emph{BLD}) {\cite{yao2020blendedmvs}} is a large-scale MVS dataset containing 113 well-reconstructed models. These scenes cover a variety of different scenes, including architectures, street-views, sculptures and small objects. 
Different from DTU, scenes in BlendedMVS contain a variety of different camera trajectories, which are more challenging.
The official split {\cite{yao2020blendedmvs}} is adopted in the experiments.

\noindent\textbf{GTA-SFM} (\emph{GTA}) {\cite{wang2020flow}} is a synthetic dataset rendered from GTA-V, an open-world game with large-scale city models. It contains 200 scenes for training and 19 scenes for testing. Various conditions like weather, daytime and indoor / outdoor are manually controlled to enlarge the diversity and usability of the dataset.
We follow the original split of train and test set used in {\cite{wang2020flow}}.
The original GTA-SFM dataset provides the intrinsic and extrinsic parameters, paired with corresponding image and depth map on that view.
The neighboring views of each view in all images are separated based on similarity among matching points following COLMAP {\cite{schonberger2016pixelwise}}.
Following the format of camera files defined in MVSNet, we further modify the camera parameters provided by GTA-SFM to support MVSNet.
Note that the extrinsic parameters in GTA-SFM is the inverse form of the ones defined in MVSNet.

\subsection{Evaluation Benchmark and Metric}

\subsubsection{DTU Benchmark}

Official MATLAB code is used for quantitative evaluation on the DTU test set.
\emph{Accuracy}, \emph{Completeness} and \emph{Overall} are used as evaluation measures:

\begin{itemize}
    \item \emph{Accuracy} is measured as the distance from the MVS reconstructed results to the ground truth point clouds, encapsulating the quality of the reconstructed MVS points;
    \item \emph{Completeness} is measured as the distance from the reference ground truth to the reconstructed results, encapsulating how much of the surface is captured by the MVS reconstruction.
    \item \emph{Overall} is the average of \emph{Accuracy} and \emph{Completeness}, which is further used for an overall assessment of the reconstruction quality.
\end{itemize}

\subsubsection{Tanks\&Temples Benchmark}

For quantitative evaluation, the reconstructed results are uploaded to the official website of \\
Tanks\&Temples Benchmark.
The test set is separated into 2 partitions: intermediate set and advanced set.
\emph{F-score} of each scene in the test set is reported on the online leaderboard.

\subsubsection{Customized Benchmark}

Since BlendedMVS and GTA-SFM datasets do not have quantitative evaluation benchmark on 3D reconstruction like DTU and Tanks\&Temples, we build a customized benchmark ourselves.
The evaluation code is implemented in PyCUDA which supports GPU parallel acceleration.

\subsubsection{Evaluation Protocol}

Following the evaluation protocol defined by Tanks\&Temples, \emph{Precision} (Eq. \ref{eqs2}), \emph{Recall} (Eq. \ref{eqs4}), and \emph{F-score} (Eq. \ref{eqs5}) are used to measure the reconstruction quality.

Assume that $G$ is the ground truth, and $R$ is the reconstructed point cloud which requires evaluation.
For each reconstructed point $r \in R$, its distance to the ground truth point cloud $G$ is defined as:
\begin{equation}
    e_{r \rightarrow G} = \min_{g \in G} \| r - g \|
    \label{eqs1}
\end{equation}
where $g$ is any point of the ground truth point cloud $G$.
Note that each point $r \in \mathbb{R}^3$ or $g \in \mathbb{R}^3$ is embedded with its X/Y/Z coordinates.

The calculated distances of all points $r$ can then be aggregated to define the precision of the reconstruction result $R$ given a threshold $d$:
\begin{equation}
    \text{Pre}(d) = \frac{100}{\|R\|} \sum_{r \in R} [ e_{r \rightarrow G} < d]
    \label{eqs2}
\end{equation}
where $[\cdot]$ is the Iverson bracket.
The precision score ($\text{Pre}(d)$) is defined to lie in the range of $[0, 100]$ and can be interpreted as a percentage.

In analogy, for each point $g$ in the ground truth reference $G$, its distance to the reconstruction result $R$ is:
\begin{equation}
    e_{g \rightarrow R} = \min_{r \in R} \| g - r \|
    \label{eqs3}
\end{equation}

Then, the recall score ($\text{Rec}(d)$) of the reconstruction $R$ can be calculated given a threshold $d$:
\begin{equation}
    \text{Rec}(d) = \frac{100}{\|G\|} \sum_{g \in G} [ e_{g \rightarrow R} < d ]
    \label{eqs4}
\end{equation}

Precision and recall scores can be further combined as a summary measure of F-score:
\begin{equation}
    F(d) = \frac{2 \text{Pre}(d) \text{Rec}(d)}{\text{Pre}(d) + \text{Rec}(d)}
    \label{eqs5}
\end{equation}

\subsubsection{Generation of Ground Truth}

For BlendedMVS, the provided ground truth mesh is comprised of many splits.
These meshes are merged into a total one first, and Monte-Carlo sampling is used to sample the point cloud uniformly from the ground truth mesh.
For a synthetic dataset like GTA-SFM, the provided test set is attached with ideal depth map ground truth.
Gipuma {\cite{galliani2015massively}} is then used to reconstruct the ideal depth maps and camera parameters into ground truth point clouds.
Note that the extrinsic matrix defined by GTA-SFM is the inverse form of the ones defined in Gipuma.

\subsection{Training}
\label{sec:exp:impl:training}

In default, we utilize CasMVSNet \cite{xiaodong2020cascade} as the backbone network.
The split of train, valid and test sets in each dataset follows the official configuration in \emph{DTU}, \emph{BLD} and \emph{GTA}.
On each dataset, the model is trained on the training set first and tested on the test set for ablation study and evaluation.
8 NVIDIA V100 GPUs are used for training.
The batch size on each GPU is set to 1 and the training procedure requires 16 epochs.
For \emph{DTU}, the resolution of all input images is set to $640 \times 512$.
For \emph{BLD}, the resolution of all input images is set to $768 \times 576$.
For \emph{GTA}, the resolution of all input images is set to $640 \times 480$.
The number of views is set to 3 following the default setting of CasMVSNet \cite{xiaodong2020cascade}.
The hyper-parameter setting of backbone networks follows the official CasMVSNet as well.
In default, Whitening and Coloring Transform (WCT \cite{li2017universal}) is selected as the backbone of STM module.
The backbone of GPM is a spatial propagation network \cite{liu2017learning}, a generic framework that can be applied to many affinity-related tasks, including but not limited to image matting, segmentation and etc.
STM and GPM are further trained on each dataset following Eq. 8 and Eq. 10 in the manuscript respectively.

\subsection{Trading Time with Space}
\label{sec:exp:impl:trading}

The generation of style-transferred samples in an online manner may occupy extra time because of forward inference in STM and GPM.
Furthermore, the neural network models defined in STM and GPM may require extra memories in GPU as well.
For simplicity, we trade time and memory with space by generating the style-transferred samples in an offline manner.
The generated images are saved in the disk via propagating the STM and GPM first.
During the training phase, these off-the-shelf images are read from the disk directly.

\subsection{Testing}
\label{sec:exp:impl:testing}

For evaluation, the model is tested on the evaluation or test set of \emph{DTU}, \emph{T\&T}, \emph{BLD}, and \emph{GTA}.
In \emph{DTU}, the resolution of input is $1152 \times 864$ and the number of views is set to 5.
In \emph{T\&T}, the resolution of input is $1920 \times 1056$ and the number of views is set to 7.
In \emph{BLD}, the resolution of input is $768 \times 576$ and the number of views is set to 5.
In \emph{GTA}, the resolution of input is $640 \times 480$ and the number of views is set to 5.


\subsection{Semi-supervised settings:}
2 different semi-supervised settings are adopted in experiments: (1) labeled and unlabeled data separated by \textbf{scenes}; (2) labeled and unlabeled data separated by \textbf{views}.
The former setting based on scenes aims at exploring the performance of our proposed method under this challenging case.
For the latter one, we aim to simulate the common case that labeled and unlabeled data are separated by the multi-view image pairs on different views.

\subsection{Backbone of STM}

A convolutional neural network (CNN) maps the input image $x \in \mathbb{R}^{H \times W \times 3}$ into a set of feature maps, $\{ F^l(x) \}_{l=1}^L$, where $F^l: \mathbb{R}^{H \times W \times 3} \rightarrow \mathbb{R}^{H \times W \times D_l}$ is the mapping from image to the $l$-th layer of CNN.
Following the assumption in \cite{gatys2015neural}, the style information is encoded by Gram Matrix $G \in \mathbb{R}^{D_l \times D_l}$, which is irrelevant to the position information but related to the abstract semantic information captured in the feature maps.
\begin{equation}
    \left[G^{l}\left(F^{l}\right)\right]_{i j}=\sum_{k} F_{i k}^{l} F_{j k}^{l}
    \label{eqs6}
\end{equation}

Given the content image $x_c$ and the style image $x_s$, the basic optimization problem of neural style transfer is to solve the following equations:
\begin{equation}
\begin{aligned}
    x^* = \text{argmin}_{x \in \mathbb{R}^{H \times W \times 3}} [\alpha L_{content}(x_c, x) + \ \beta L_{style}(x_s, x)]
    \label{eqs7}
\end{aligned}
\end{equation}
with 
\begin{equation}
    L_{content} = \| F^l(x_c) - F^l(x) \|_2^2
    \label{eqs8}
\end{equation}
\begin{equation}
    L_{style} = \sum_{l=1}^L \| G^l[F^l(x_s)] - G^l[F^l(x)] \|_2^2
    \label{eqs9}
\end{equation}

By solving  Eq. \ref{eqs7} and utilizing the VGG network as the CNN backbone, the basic neural style transfer algorithm can be implemented in the STM.
As a faster neural style transfer algorithm, Whitening and Coloring Transform (WCT \cite{li2017universal}) is selected to as the backbone STM module.
The basic concepts of WCT are already introduced in Section 3.7 of the manuscript.
We borrow the popular implementation of WCT\footnote{\url{https://github.com/sunshineatnoon/PytorchWCT}} in STM.
The autoencoder of WCT is trained renewedly on the images of MVS datasets with the losses discussed in the manuscript.

\subsubsection{Backbone of GPM}

\begin{figure*}[t] 
\centering 
\includegraphics[width=\hsize]{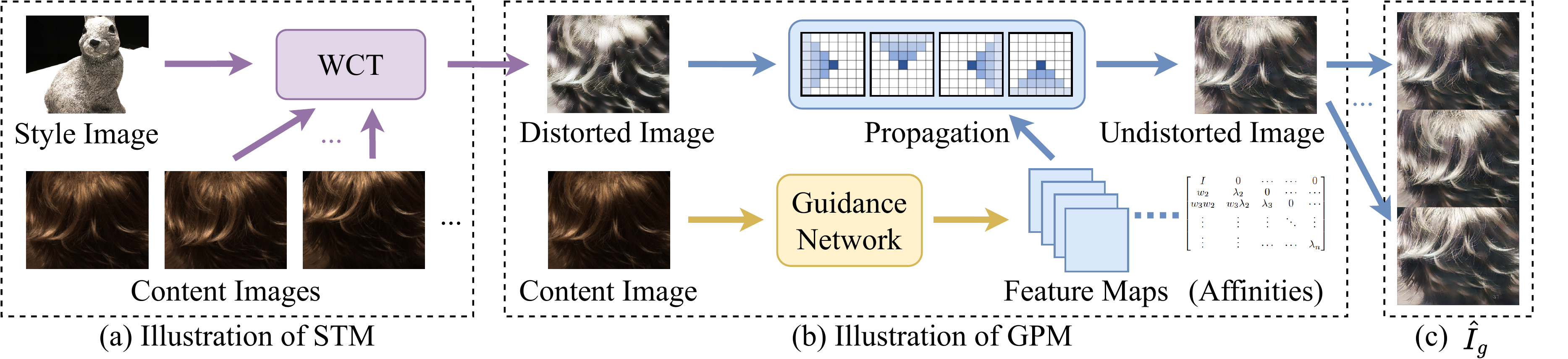} 
\caption{Illustration of STM and GPM.} 
\label{fig_illustration_stm_gpm} 
\end{figure*}

\begin{figure*}[t]
\centering
\includegraphics[width=0.9\hsize]{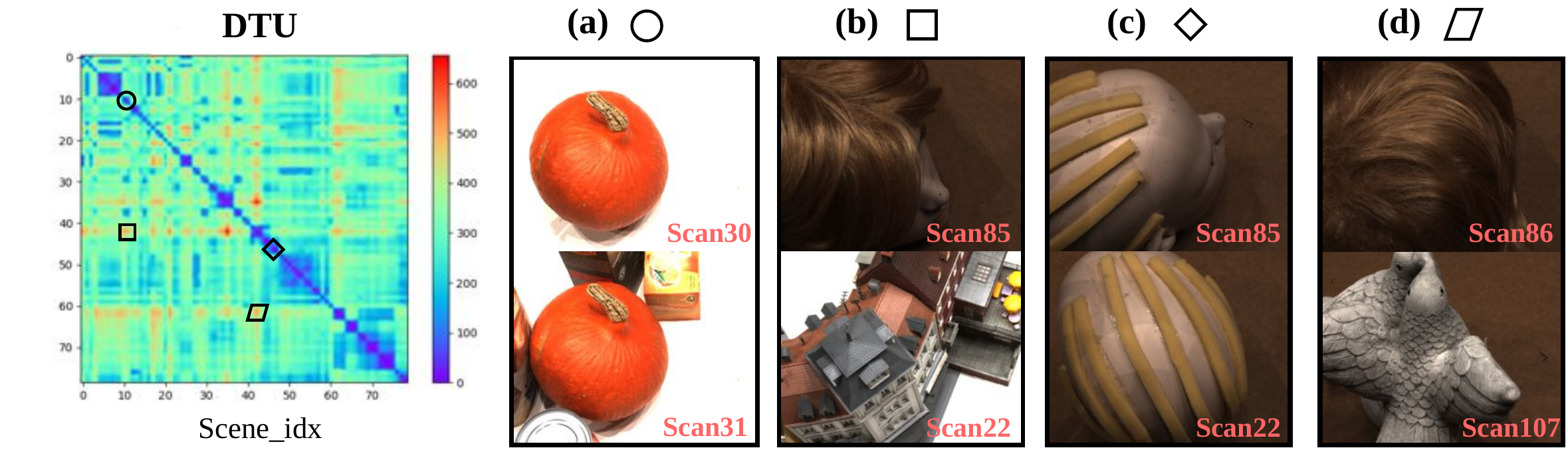} 
\caption{Some examples of visualized distribution distance among views on DTU dataset.}
\label{fig:example_distribution}
\end{figure*}

The backbone of GPM is a spatial propagation network \cite{liu2017learning}, a generic framework\footnote{\url{https://github.com/Liusifei/pytorch_spn}} that can be applied to many affinity-related tasks, including but not limited to image matting, segmentation and etc.
A brief introduction of the STM and GPM is illustrated in Fig. \ref{fig_illustration_stm_gpm}.
In the forward propagation, the labeled and unlabeled images are selected as content and style images respectively.
The output of STM is the style transferred image which has unexpected distortions that may lose 3D geometric details.
In GPM, the original content image is fed to a Guidance Network constructed by a CNN, and the output feature maps with customized forms that can model the affinities of local regions.
The theoretical proof is also presented in \cite{liu2017learning}.
Then, the spatial propagation operation acts as a refining process to remove the distortions in the style transferred image.
The generated images $\hat{I}_g$ is then fed to the key loss discussed in Eq. 12 of the manuscript.

\section{Supplementary Results and Discussions}

\subsection{Distribution Distance among Scenes}

In Figure 1 of the manuscript, we provide a brief illustration of distribution gap among scenes.
For each scene, each of the multi-view images is fed to a ImageNet-pretrained VGG network to obtain a feature embedding.
The feature embeddings of all views are concatenated as a set describing the distribution of a scene.
To measure the distribution gap among scenes, the standard Maximum Mean Discrepancy (MMD) distance \cite{muandet2017kernel} is adopted in our implementation.
We further visualize some examples of distribution distance among scenes in Figure \ref{fig:example_distribution}.

\subsection{Visualization of 3D Consistency Test}

\begin{figure*}[t] 
\centering 
\includegraphics[width=\textwidth]{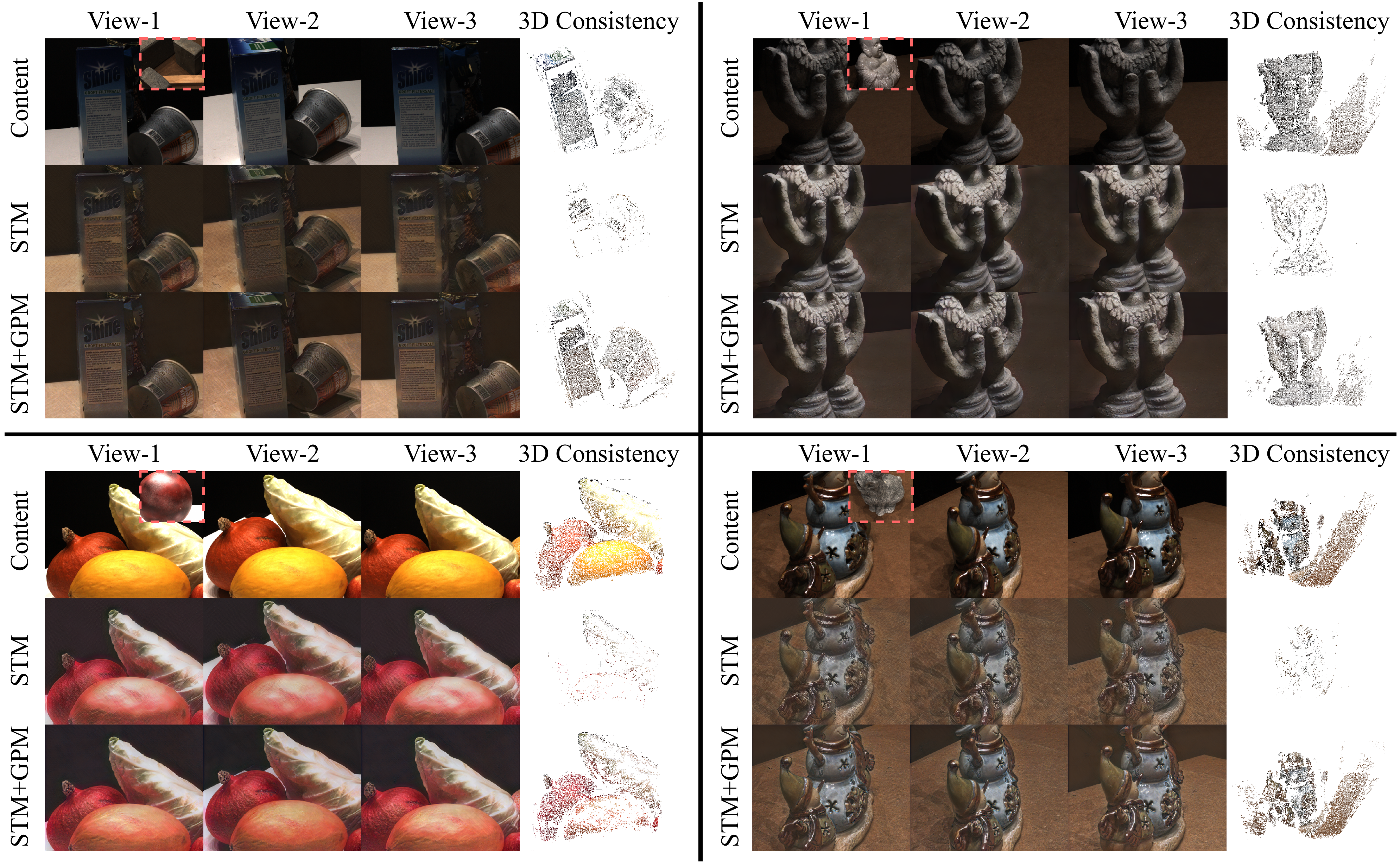} 
\caption{Visualization of 3D consistency via running COLMAP \cite{schonberger2016pixelwise}.} 
\label{fig_style_transfer_results} 
\end{figure*}

As an extension to Fig. 2 in the manuscript, we also provide more examples of style transfer results as well as the 3D consistency visualization.
The 3D consistency test is conducted via running COLMAP \cite{schonberger2016pixelwise} on the multi-view images.
The reconstructed point clouds and the multi-view images are visualized in Fig. \ref{fig_style_transfer_results}.
From the comparison between original STM and STM+GPM, we can find that the 3D  details after style transfer can successfully preserved with the GPM module.

\subsection{Visualization of Reconstruction Results}

\begin{figure*}[t] 
\centering 
\includegraphics[width=\textwidth]{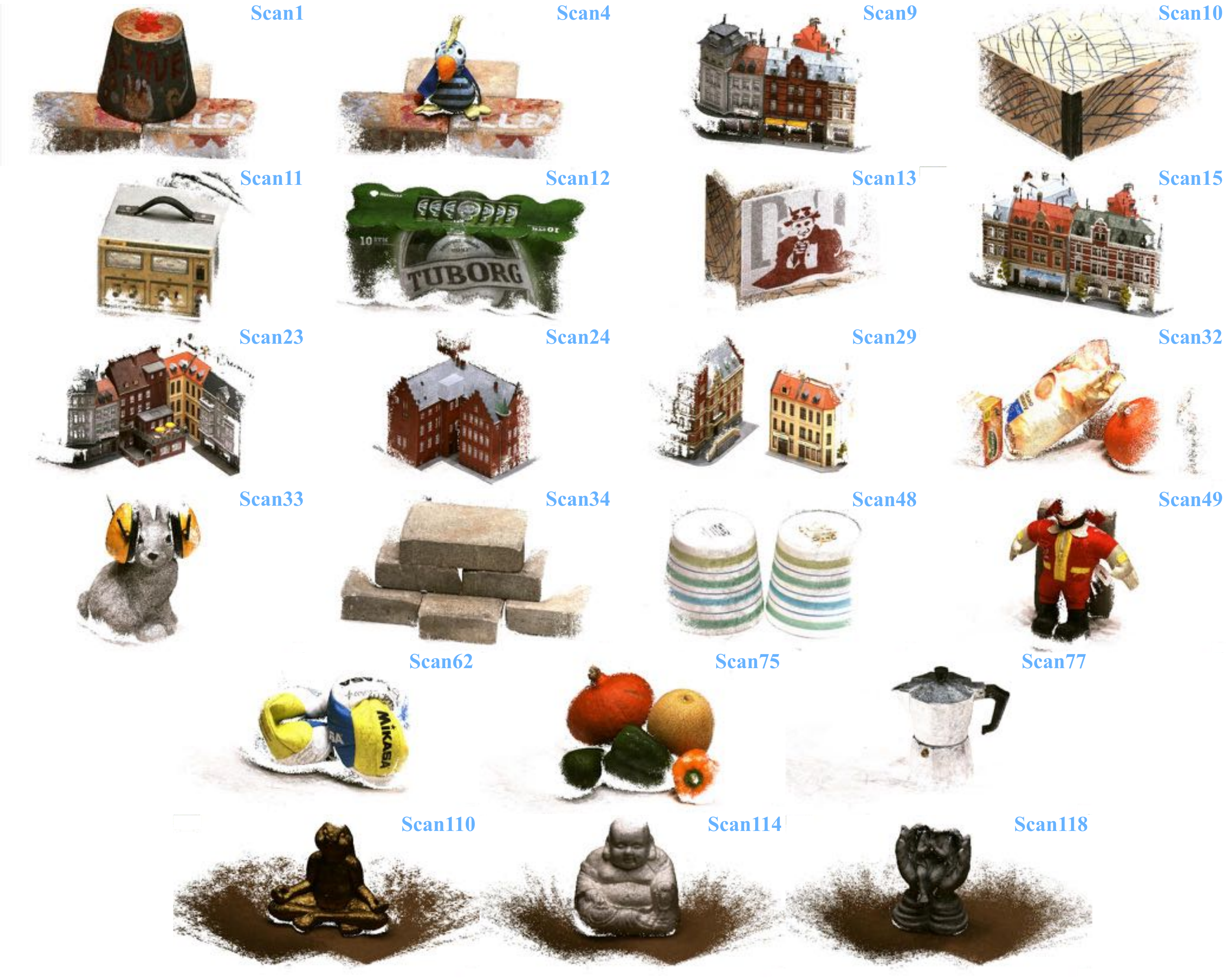} 
\caption{Visualization of all scenes in the test set of DTU dataset.} 
\label{fig_dtu_visualization_all} 
\end{figure*}

\begin{figure*}[t] 
\centering 
\includegraphics[width=\textwidth]{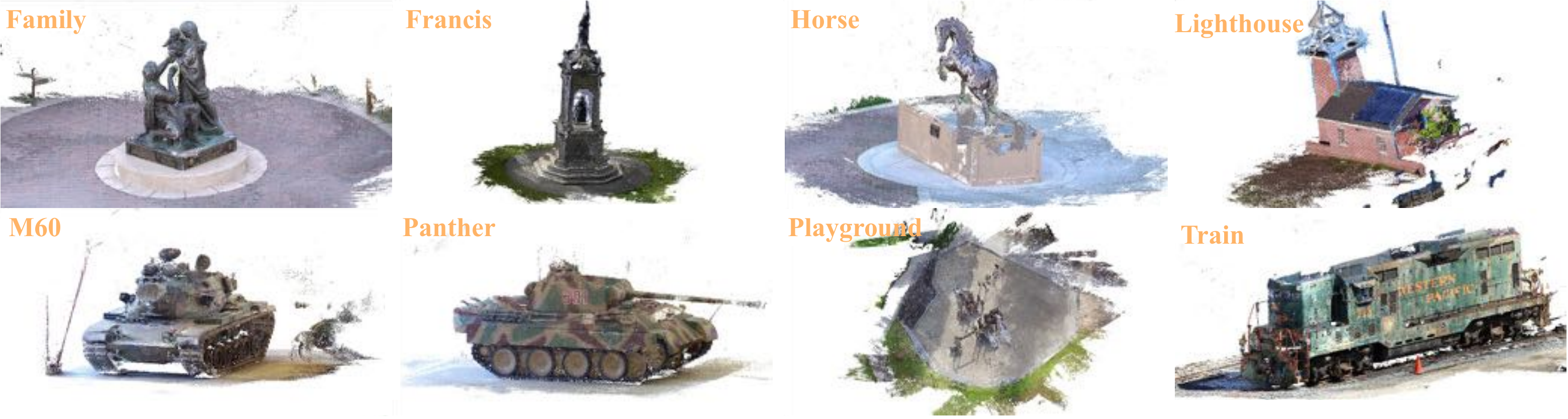} 
\caption{Visualization of all scenes in the intermediate partition of Tanks\&Temples dataset.} 
\label{fig_tt_intermeidate_visualziation} 
\end{figure*}

\begin{figure*}[t] 
\centering 
\includegraphics[width=\textwidth]{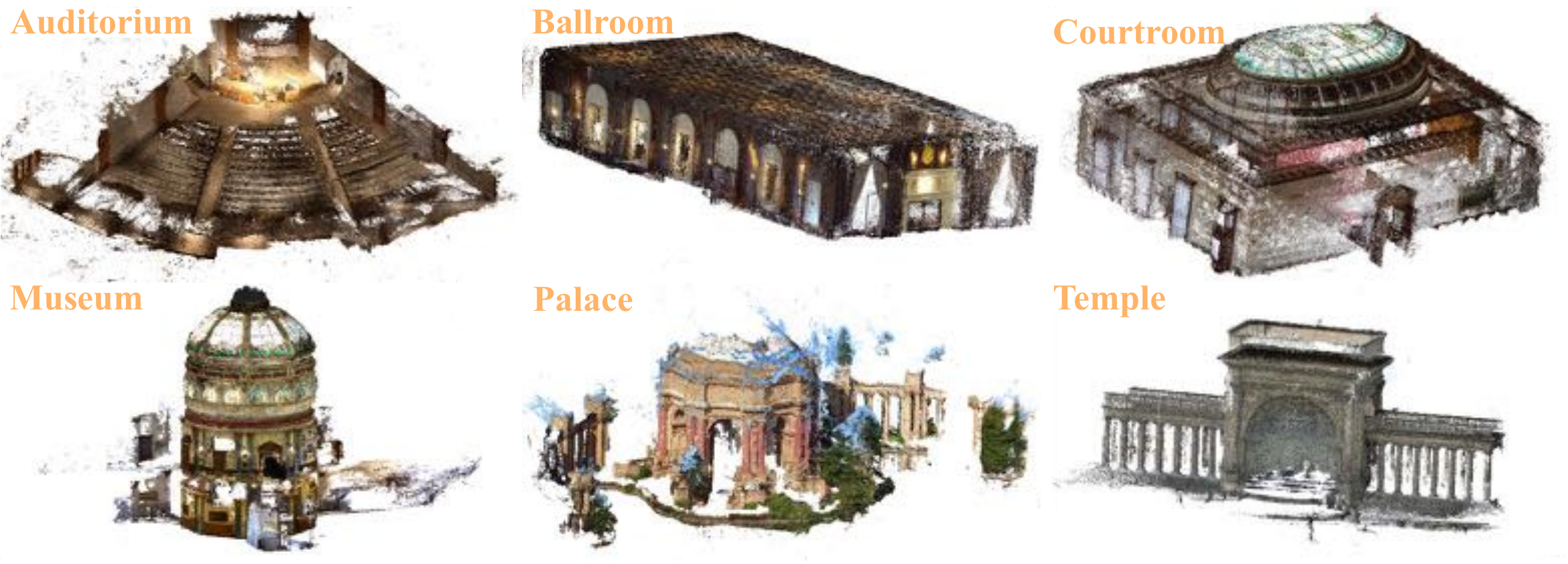} 
\caption{Visualization of all scenes in the advanced partition of Tanks\&Temples dataset.} 
\label{fig_tt_advanced_visualization} 
\end{figure*}

\begin{figure*}[t] 
\centering 
\includegraphics[width=\textwidth]{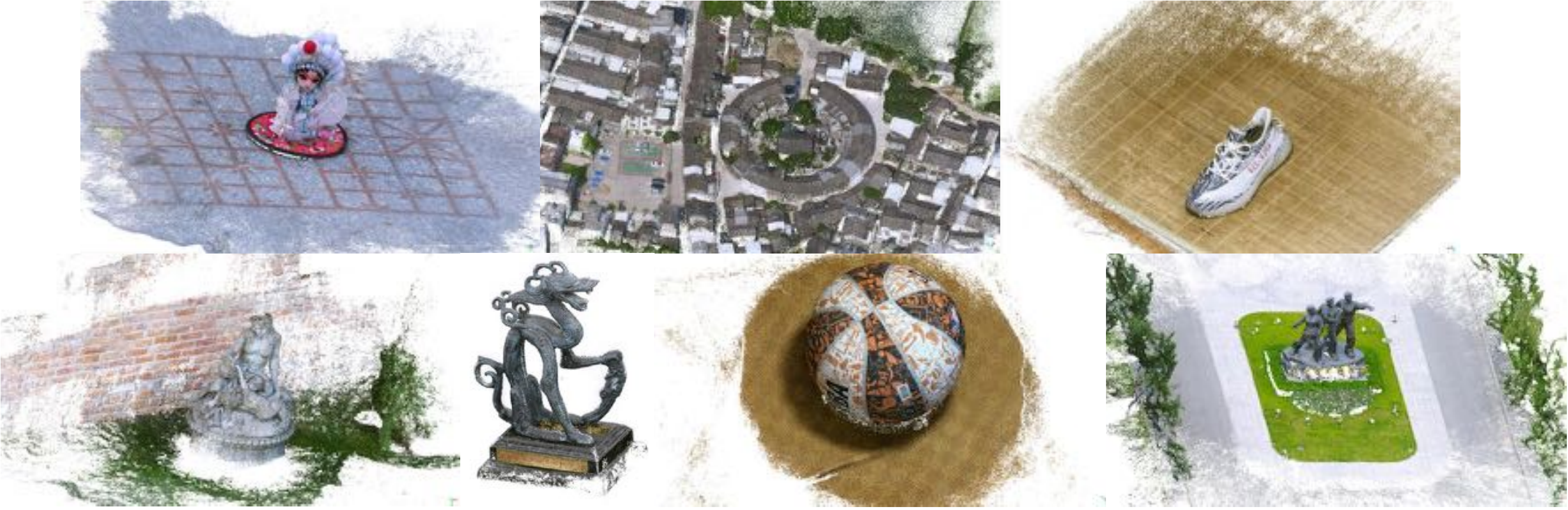} 
\caption{Visualization of all scenes in the test set of BlendedMVS dataset.} 
\label{fig_blendedmvs_visualization} 
\end{figure*}

\begin{figure*}[t] 
\centering 
\includegraphics[width=\textwidth]{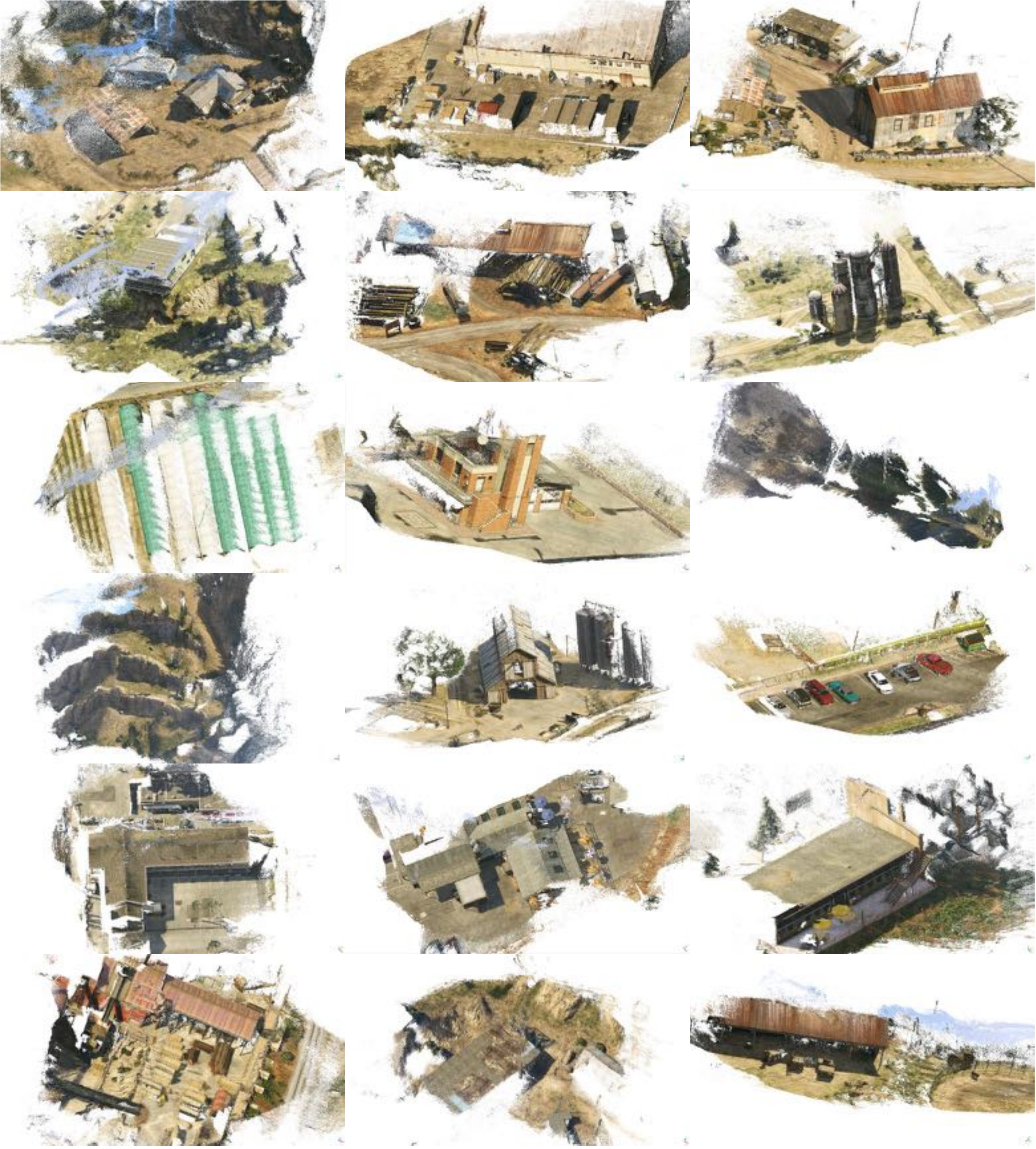} 
\caption{Visualization of all scenes in the test set of GTASFM dataset.} 
\label{fig_gtasfm_visualization} 
\end{figure*}

By running the whole procedure of 3D reconstruction, we can obtain the point clouds from the multi-view images in the test set of each MVS dataset.
In Fig. \ref{fig_dtu_visualization_all}, we present the visualization results of all scenes in the test set of DTU benchmark.
And the reconstructed point clouds of the intermediate and advanced partition in Tanks\&Temples dataset are shown in Fig. \ref{fig_tt_intermeidate_visualziation} and \ref{fig_tt_advanced_visualization}.
The visualized point clouds of BlendedMVS and GTA-SFM are further presented in Fig. \ref{fig_blendedmvs_visualization} and \ref{fig_gtasfm_visualization}.

\bibliographystyle{ACM-Reference-Format}
\balance
\bibliography{reference}


\begin{thebibliography}{54}


\ifx \showCODEN    \undefined \def \showCODEN     #1{\unskip}     \fi
\ifx \showDOI      \undefined \def \showDOI       #1{#1}\fi
\ifx \showISBNx    \undefined \def \showISBNx     #1{\unskip}     \fi
\ifx \showISBNxiii \undefined \def \showISBNxiii  #1{\unskip}     \fi
\ifx \showISSN     \undefined \def \showISSN      #1{\unskip}     \fi
\ifx \showLCCN     \undefined \def \showLCCN      #1{\unskip}     \fi
\ifx \shownote     \undefined \def \shownote      #1{#1}          \fi
\ifx \showarticletitle \undefined \def \showarticletitle #1{#1}   \fi
\ifx \showURL      \undefined \def \showURL       {\relax}        \fi
\providecommand\bibfield[2]{#2}
\providecommand\bibinfo[2]{#2}
\providecommand\natexlab[1]{#1}
\providecommand\showeprint[2][]{arXiv:#2}

\bibitem[Bachman et~al\mbox{.}(2014)]%
        {bachman2014learning}
\bibfield{author}{\bibinfo{person}{Philip Bachman}, \bibinfo{person}{Ouais
  Alsharif}, {and} \bibinfo{person}{Doina Precup}.}
  \bibinfo{year}{2014}\natexlab{}.
\newblock \showarticletitle{Learning with pseudo-ensembles}.
\newblock \bibinfo{journal}{\emph{Advances in neural information processing
  systems}}  \bibinfo{volume}{27} (\bibinfo{year}{2014}).
\newblock


\bibitem[Berthelot et~al\mbox{.}(2019)]%
        {berthelot2019mixmatch}
\bibfield{author}{\bibinfo{person}{David Berthelot}, \bibinfo{person}{Nicholas
  Carlini}, \bibinfo{person}{Ian Goodfellow}, \bibinfo{person}{Nicolas
  Papernot}, \bibinfo{person}{Avital Oliver}, {and} \bibinfo{person}{Colin~A
  Raffel}.} \bibinfo{year}{2019}\natexlab{}.
\newblock \showarticletitle{Mixmatch: A holistic approach to semi-supervised
  learning}.
\newblock \bibinfo{journal}{\emph{Advances in Neural Information Processing
  Systems}}  \bibinfo{volume}{32} (\bibinfo{year}{2019}).
\newblock


\bibitem[Campbell et~al\mbox{.}(2008)]%
        {campbell2008using}
\bibfield{author}{\bibinfo{person}{Neill~DF Campbell}, \bibinfo{person}{George
  Vogiatzis}, \bibinfo{person}{Carlos Hern{\'a}ndez}, {and}
  \bibinfo{person}{Roberto Cipolla}.} \bibinfo{year}{2008}\natexlab{}.
\newblock \showarticletitle{Using multiple hypotheses to improve depth-maps for
  multi-view stereo}. In \bibinfo{booktitle}{\emph{European Conference on
  Computer Vision}}. Springer, \bibinfo{pages}{766--779}.
\newblock


\bibitem[Chen et~al\mbox{.}(2020)]%
        {chen2020mvsnet++}
\bibfield{author}{\bibinfo{person}{Po-Heng Chen}, \bibinfo{person}{Hsiao-Chien
  Yang}, \bibinfo{person}{Kuan-Wen Chen}, {and} \bibinfo{person}{Yong-Sheng
  Chen}.} \bibinfo{year}{2020}\natexlab{}.
\newblock \showarticletitle{Mvsnet++: Learning depth-based attention pyramid
  features for multi-view stereo}.
\newblock \bibinfo{journal}{\emph{IEEE Transactions on Image Processing}}
  \bibinfo{volume}{29} (\bibinfo{year}{2020}), \bibinfo{pages}{7261--7273}.
\newblock


\bibitem[Chen et~al\mbox{.}(2019)]%
        {chen2019point}
\bibfield{author}{\bibinfo{person}{Rui Chen}, \bibinfo{person}{Songfang Han},
  \bibinfo{person}{Jing Xu}, {and} \bibinfo{person}{Hao Su}.}
  \bibinfo{year}{2019}\natexlab{}.
\newblock \showarticletitle{Point-based multi-view stereo network}. In
  \bibinfo{booktitle}{\emph{Proceedings of the IEEE/CVF International
  Conference on Computer Vision}}. \bibinfo{pages}{1538--1547}.
\newblock


\bibitem[Cheng et~al\mbox{.}(2020)]%
        {cheng2020deep}
\bibfield{author}{\bibinfo{person}{Shuo Cheng}, \bibinfo{person}{Zexiang Xu},
  \bibinfo{person}{Shilin Zhu}, \bibinfo{person}{Zhuwen Li},
  \bibinfo{person}{Li~Erran Li}, \bibinfo{person}{Ravi Ramamoorthi}, {and}
  \bibinfo{person}{Hao Su}.} \bibinfo{year}{2020}\natexlab{}.
\newblock \showarticletitle{Deep stereo using adaptive thin volume
  representation with uncertainty awareness}. In
  \bibinfo{booktitle}{\emph{Proceedings of the IEEE/CVF Conference on Computer
  Vision and Pattern Recognition}}. \bibinfo{pages}{2524--2534}.
\newblock


\bibitem[Dai et~al\mbox{.}(2019)]%
        {dai2019mvs2}
\bibfield{author}{\bibinfo{person}{Yuchao Dai}, \bibinfo{person}{Zhidong Zhu},
  \bibinfo{person}{Zhibo Rao}, {and} \bibinfo{person}{Bo Li}.}
  \bibinfo{year}{2019}\natexlab{}.
\newblock \showarticletitle{Mvs2: Deep unsupervised multi-view stereo with
  multi-view symmetry}. In \bibinfo{booktitle}{\emph{2019 International
  Conference on 3D Vision (3DV)}}. IEEE, \bibinfo{pages}{1--8}.
\newblock


\bibitem[Fuhrmann et~al\mbox{.}(2014)]%
        {fuhrmann2014mve}
\bibfield{author}{\bibinfo{person}{Simon Fuhrmann}, \bibinfo{person}{Fabian
  Langguth}, {and} \bibinfo{person}{Michael Goesele}.}
  \bibinfo{year}{2014}\natexlab{}.
\newblock \showarticletitle{Mve-a multi-view reconstruction environment.}. In
  \bibinfo{booktitle}{\emph{GCH}}. \bibinfo{pages}{11--18}.
\newblock


\bibitem[Furukawa and Ponce(2009)]%
        {furukawa2009accurate}
\bibfield{author}{\bibinfo{person}{Yasutaka Furukawa} {and}
  \bibinfo{person}{Jean Ponce}.} \bibinfo{year}{2009}\natexlab{}.
\newblock \showarticletitle{Accurate, dense, and robust multiview stereopsis}.
\newblock \bibinfo{journal}{\emph{IEEE transactions on pattern analysis and
  machine intelligence}} \bibinfo{volume}{32}, \bibinfo{number}{8}
  (\bibinfo{year}{2009}), \bibinfo{pages}{1362--1376}.
\newblock


\bibitem[Galliani et~al\mbox{.}(2015)]%
        {galliani2015massively}
\bibfield{author}{\bibinfo{person}{Silvano Galliani}, \bibinfo{person}{Katrin
  Lasinger}, {and} \bibinfo{person}{Konrad Schindler}.}
  \bibinfo{year}{2015}\natexlab{}.
\newblock \showarticletitle{Massively parallel multiview stereopsis by surface
  normal diffusion}. In \bibinfo{booktitle}{\emph{Proceedings of the IEEE
  International Conference on Computer Vision}}. \bibinfo{pages}{873--881}.
\newblock


\bibitem[Gatys et~al\mbox{.}(2015)]%
        {gatys2015neural}
\bibfield{author}{\bibinfo{person}{Leon~A Gatys}, \bibinfo{person}{Alexander~S
  Ecker}, {and} \bibinfo{person}{Matthias Bethge}.}
  \bibinfo{year}{2015}\natexlab{}.
\newblock \showarticletitle{A neural algorithm of artistic style}.
\newblock \bibinfo{journal}{\emph{arXiv preprint arXiv:1508.06576}}
  (\bibinfo{year}{2015}).
\newblock


\bibitem[Grandvalet and Bengio(2004)]%
        {grandvalet2004semi}
\bibfield{author}{\bibinfo{person}{Yves Grandvalet} {and}
  \bibinfo{person}{Yoshua Bengio}.} \bibinfo{year}{2004}\natexlab{}.
\newblock \showarticletitle{Semi-supervised learning by entropy minimization}.
\newblock \bibinfo{journal}{\emph{Advances in neural information processing
  systems}}  \bibinfo{volume}{17} (\bibinfo{year}{2004}).
\newblock


\bibitem[Huang et~al\mbox{.}(2021)]%
        {huang2021m3vsnet}
\bibfield{author}{\bibinfo{person}{Baichuan Huang}, \bibinfo{person}{Hongwei
  Yi}, \bibinfo{person}{Can Huang}, \bibinfo{person}{Yijia He},
  \bibinfo{person}{Jingbin Liu}, {and} \bibinfo{person}{Xiao Liu}.}
  \bibinfo{year}{2021}\natexlab{}.
\newblock \showarticletitle{Mˆ3VSNet: Unsupervised multi-metric multi-view
  stereo network}. In \bibinfo{booktitle}{\emph{2021 IEEE International
  Conference on Image Processing (ICIP)}}. IEEE, \bibinfo{pages}{3163--3167}.
\newblock


\bibitem[Jensen et~al\mbox{.}(2014)]%
        {jensen2014large}
\bibfield{author}{\bibinfo{person}{Rasmus Jensen}, \bibinfo{person}{Anders
  Dahl}, \bibinfo{person}{George Vogiatzis}, \bibinfo{person}{Engin Tola},
  {and} \bibinfo{person}{Henrik Aan{\ae}s}.} \bibinfo{year}{2014}\natexlab{}.
\newblock \showarticletitle{Large scale multi-view stereopsis evaluation}. In
  \bibinfo{booktitle}{\emph{Proceedings of the IEEE conference on computer
  vision and pattern recognition}}. \bibinfo{pages}{406--413}.
\newblock


\bibitem[Ji et~al\mbox{.}(2017)]%
        {ji2017surfacenet}
\bibfield{author}{\bibinfo{person}{Mengqi Ji}, \bibinfo{person}{Juergen Gall},
  \bibinfo{person}{Haitian Zheng}, \bibinfo{person}{Yebin Liu}, {and}
  \bibinfo{person}{Lu Fang}.} \bibinfo{year}{2017}\natexlab{}.
\newblock \showarticletitle{Surfacenet: An end-to-end 3d neural network for
  multiview stereopsis}. In \bibinfo{booktitle}{\emph{Proceedings of the IEEE
  International Conference on Computer Vision}}. \bibinfo{pages}{2307--2315}.
\newblock


\bibitem[Kendall et~al\mbox{.}(2017)]%
        {kendall2017end}
\bibfield{author}{\bibinfo{person}{Alex Kendall}, \bibinfo{person}{Hayk
  Martirosyan}, \bibinfo{person}{Saumitro Dasgupta}, \bibinfo{person}{Peter
  Henry}, \bibinfo{person}{Ryan Kennedy}, \bibinfo{person}{Abraham Bachrach},
  {and} \bibinfo{person}{Adam Bry}.} \bibinfo{year}{2017}\natexlab{}.
\newblock \showarticletitle{End-to-end learning of geometry and context for
  deep stereo regression}. In \bibinfo{booktitle}{\emph{Proceedings of the IEEE
  international conference on computer vision}}. \bibinfo{pages}{66--75}.
\newblock


\bibitem[Khot et~al\mbox{.}(2019)]%
        {khot2019learning}
\bibfield{author}{\bibinfo{person}{Tejas Khot}, \bibinfo{person}{Shubham
  Agrawal}, \bibinfo{person}{Shubham Tulsiani}, \bibinfo{person}{Christoph
  Mertz}, \bibinfo{person}{Simon Lucey}, {and} \bibinfo{person}{Martial
  Hebert}.} \bibinfo{year}{2019}\natexlab{}.
\newblock \showarticletitle{Learning unsupervised multi-view stereopsis via
  robust photometric consistency}.
\newblock \bibinfo{journal}{\emph{arXiv preprint arXiv:1905.02706}}
  (\bibinfo{year}{2019}).
\newblock


\bibitem[Kim et~al\mbox{.}(2021)]%
        {kim2021just}
\bibfield{author}{\bibinfo{person}{Taekyung Kim}, \bibinfo{person}{Jaehoon
  Choi}, \bibinfo{person}{Seokeon Choi}, \bibinfo{person}{Dongki Jung}, {and}
  \bibinfo{person}{Changick Kim}.} \bibinfo{year}{2021}\natexlab{}.
\newblock \showarticletitle{Just a Few Points Are All You Need for Multi-View
  Stereo: A Novel Semi-Supervised Learning Method for Multi-View Stereo}. In
  \bibinfo{booktitle}{\emph{Proceedings of the IEEE/CVF International
  Conference on Computer Vision}}. \bibinfo{pages}{6178--6186}.
\newblock


\bibitem[Knapitsch et~al\mbox{.}(2017)]%
        {knapitsch2017tanks}
\bibfield{author}{\bibinfo{person}{Arno Knapitsch}, \bibinfo{person}{Jaesik
  Park}, \bibinfo{person}{Qian-Yi Zhou}, {and} \bibinfo{person}{Vladlen
  Koltun}.} \bibinfo{year}{2017}\natexlab{}.
\newblock \showarticletitle{Tanks and temples: Benchmarking large-scale scene
  reconstruction}.
\newblock \bibinfo{journal}{\emph{ACM Transactions on Graphics (ToG)}}
  \bibinfo{volume}{36}, \bibinfo{number}{4} (\bibinfo{year}{2017}),
  \bibinfo{pages}{1--13}.
\newblock


\bibitem[Langguth et~al\mbox{.}(2016)]%
        {langguth2016shading}
\bibfield{author}{\bibinfo{person}{Fabian Langguth}, \bibinfo{person}{Kalyan
  Sunkavalli}, \bibinfo{person}{Sunil Hadap}, {and} \bibinfo{person}{Michael
  Goesele}.} \bibinfo{year}{2016}\natexlab{}.
\newblock \showarticletitle{Shading-aware multi-view stereo}. In
  \bibinfo{booktitle}{\emph{European Conference on Computer Vision}}. Springer,
  \bibinfo{pages}{469--485}.
\newblock


\bibitem[Li et~al\mbox{.}(2017)]%
        {li2017universal}
\bibfield{author}{\bibinfo{person}{Yijun Li}, \bibinfo{person}{Chen Fang},
  \bibinfo{person}{Jimei Yang}, \bibinfo{person}{Zhaowen Wang},
  \bibinfo{person}{Xin Lu}, {and} \bibinfo{person}{Ming-Hsuan Yang}.}
  \bibinfo{year}{2017}\natexlab{}.
\newblock \showarticletitle{Universal style transfer via feature transforms}.
\newblock \bibinfo{journal}{\emph{Advances in neural information processing
  systems}}  \bibinfo{volume}{30} (\bibinfo{year}{2017}).
\newblock


\bibitem[Liu et~al\mbox{.}(2017)]%
        {liu2017learning}
\bibfield{author}{\bibinfo{person}{Sifei Liu}, \bibinfo{person}{Shalini
  De~Mello}, \bibinfo{person}{Jinwei Gu}, \bibinfo{person}{Guangyu Zhong},
  \bibinfo{person}{Ming-Hsuan Yang}, {and} \bibinfo{person}{Jan Kautz}.}
  \bibinfo{year}{2017}\natexlab{}.
\newblock \showarticletitle{Learning affinity via spatial propagation
  networks}.
\newblock \bibinfo{journal}{\emph{Advances in Neural Information Processing
  Systems}}  \bibinfo{volume}{30} (\bibinfo{year}{2017}).
\newblock


\bibitem[Luo et~al\mbox{.}(2019)]%
        {luo2019p}
\bibfield{author}{\bibinfo{person}{Keyang Luo}, \bibinfo{person}{Tao Guan},
  \bibinfo{person}{Lili Ju}, \bibinfo{person}{Haipeng Huang}, {and}
  \bibinfo{person}{Yawei Luo}.} \bibinfo{year}{2019}\natexlab{}.
\newblock \showarticletitle{P-mvsnet: Learning patch-wise matching confidence
  aggregation for multi-view stereo}. In \bibinfo{booktitle}{\emph{Proceedings
  of the IEEE/CVF International Conference on Computer Vision}}.
  \bibinfo{pages}{10452--10461}.
\newblock


\bibitem[Ma et~al\mbox{.}(2021)]%
        {ma2021epp}
\bibfield{author}{\bibinfo{person}{Xinjun Ma}, \bibinfo{person}{Yue Gong},
  \bibinfo{person}{Qirui Wang}, \bibinfo{person}{Jingwei Huang},
  \bibinfo{person}{Lei Chen}, {and} \bibinfo{person}{Fan Yu}.}
  \bibinfo{year}{2021}\natexlab{}.
\newblock \showarticletitle{EPP-MVSNet: Epipolar-Assembling Based Depth
  Prediction for Multi-View Stereo}. In \bibinfo{booktitle}{\emph{Proceedings
  of the IEEE/CVF International Conference on Computer Vision}}.
  \bibinfo{pages}{5732--5740}.
\newblock


\bibitem[Mallick et~al\mbox{.}(2020)]%
        {mallick2020learning}
\bibfield{author}{\bibinfo{person}{Arijit Mallick}, \bibinfo{person}{J{\"o}rg
  St{\"u}ckler}, {and} \bibinfo{person}{Hendrik Lensch}.}
  \bibinfo{year}{2020}\natexlab{}.
\newblock \showarticletitle{Learning to adapt multi-view stereo by
  self-supervision}.
\newblock \bibinfo{journal}{\emph{arXiv preprint arXiv:2009.13278}}
  (\bibinfo{year}{2020}).
\newblock


\bibitem[Miyato et~al\mbox{.}(2018)]%
        {miyato2018virtual}
\bibfield{author}{\bibinfo{person}{Takeru Miyato}, \bibinfo{person}{Shin-ichi
  Maeda}, \bibinfo{person}{Masanori Koyama}, {and} \bibinfo{person}{Shin
  Ishii}.} \bibinfo{year}{2018}\natexlab{}.
\newblock \showarticletitle{Virtual adversarial training: a regularization
  method for supervised and semi-supervised learning}.
\newblock \bibinfo{journal}{\emph{IEEE transactions on pattern analysis and
  machine intelligence}} \bibinfo{volume}{41}, \bibinfo{number}{8}
  (\bibinfo{year}{2018}), \bibinfo{pages}{1979--1993}.
\newblock


\bibitem[Muandet et~al\mbox{.}(2017)]%
        {muandet2017kernel}
\bibfield{author}{\bibinfo{person}{Krikamol Muandet}, \bibinfo{person}{Kenji
  Fukumizu}, \bibinfo{person}{Bharath Sriperumbudur}, \bibinfo{person}{Bernhard
  Sch{\"o}lkopf}, {et~al\mbox{.}}} \bibinfo{year}{2017}\natexlab{}.
\newblock \showarticletitle{Kernel mean embedding of distributions: A review
  and beyond}.
\newblock \bibinfo{journal}{\emph{Foundations and Trends{\textregistered} in
  Machine Learning}} \bibinfo{volume}{10}, \bibinfo{number}{1-2}
  (\bibinfo{year}{2017}), \bibinfo{pages}{1--141}.
\newblock


\bibitem[Nievergelt and Preparata(1982)]%
        {nievergelt1982plane}
\bibfield{author}{\bibinfo{person}{J{\"u}rg Nievergelt} {and}
  \bibinfo{person}{Franco~P Preparata}.} \bibinfo{year}{1982}\natexlab{}.
\newblock \showarticletitle{Plane-sweep algorithms for intersecting geometric
  figures}.
\newblock \bibinfo{journal}{\emph{Commun. ACM}} \bibinfo{volume}{25},
  \bibinfo{number}{10} (\bibinfo{year}{1982}), \bibinfo{pages}{739--747}.
\newblock


\bibitem[Rasmus et~al\mbox{.}(2015)]%
        {rasmus2015semi}
\bibfield{author}{\bibinfo{person}{Antti Rasmus}, \bibinfo{person}{Mathias
  Berglund}, \bibinfo{person}{Mikko Honkala}, \bibinfo{person}{Harri Valpola},
  {and} \bibinfo{person}{Tapani Raiko}.} \bibinfo{year}{2015}\natexlab{}.
\newblock \showarticletitle{Semi-supervised learning with ladder networks}.
\newblock \bibinfo{journal}{\emph{Advances in neural information processing
  systems}}  \bibinfo{volume}{28} (\bibinfo{year}{2015}).
\newblock


\bibitem[Sajjadi et~al\mbox{.}(2016)]%
        {sajjadi2016regularization}
\bibfield{author}{\bibinfo{person}{Mehdi Sajjadi}, \bibinfo{person}{Mehran
  Javanmardi}, {and} \bibinfo{person}{Tolga Tasdizen}.}
  \bibinfo{year}{2016}\natexlab{}.
\newblock \showarticletitle{Regularization with stochastic transformations and
  perturbations for deep semi-supervised learning}.
\newblock \bibinfo{journal}{\emph{Advances in neural information processing
  systems}}  \bibinfo{volume}{29} (\bibinfo{year}{2016}).
\newblock


\bibitem[Samuli and Timo(2017)]%
        {samuli2017temporal}
\bibfield{author}{\bibinfo{person}{Laine Samuli} {and} \bibinfo{person}{Aila
  Timo}.} \bibinfo{year}{2017}\natexlab{}.
\newblock \showarticletitle{Temporal ensembling for semi-supervised learning}.
  In \bibinfo{booktitle}{\emph{International Conference on Learning
  Representations (ICLR)}}, Vol.~\bibinfo{volume}{4}. \bibinfo{pages}{6}.
\newblock


\bibitem[Sch{\"o}nberger et~al\mbox{.}(2016)]%
        {schonberger2016pixelwise}
\bibfield{author}{\bibinfo{person}{Johannes~L Sch{\"o}nberger},
  \bibinfo{person}{Enliang Zheng}, \bibinfo{person}{Jan-Michael Frahm}, {and}
  \bibinfo{person}{Marc Pollefeys}.} \bibinfo{year}{2016}\natexlab{}.
\newblock \showarticletitle{Pixelwise view selection for unstructured
  multi-view stereo}. In \bibinfo{booktitle}{\emph{European Conference on
  Computer Vision}}. Springer, \bibinfo{pages}{501--518}.
\newblock


\bibitem[Simonyan and Zisserman(2015)]%
        {vgg}
\bibfield{author}{\bibinfo{person}{K. Simonyan} {and} \bibinfo{person}{A.
  Zisserman}.} \bibinfo{year}{2015}\natexlab{}.
\newblock \showarticletitle{Very Deep Convolutional Networks for Large-Scale
  Image Recognition}. In \bibinfo{booktitle}{\emph{International Conference on
  Learning Representations}}.
\newblock


\bibitem[Su et~al\mbox{.}(2022)]%
        {su2022uncertainty}
\bibfield{author}{\bibinfo{person}{Wanjuan Su}, \bibinfo{person}{Qingshan Xu},
  {and} \bibinfo{person}{Wenbing Tao}.} \bibinfo{year}{2022}\natexlab{}.
\newblock \showarticletitle{Uncertainty Guided Multi-View Stereo Network for
  Depth Estimation}.
\newblock \bibinfo{journal}{\emph{IEEE Transactions on Circuits and Systems for
  Video Technology}} \bibinfo{volume}{32}, \bibinfo{number}{11}
  (\bibinfo{year}{2022}), \bibinfo{pages}{7796--7808}.
\newblock


\bibitem[Tarvainen and Valpola(2017)]%
        {tarvainen2017mean}
\bibfield{author}{\bibinfo{person}{Antti Tarvainen} {and}
  \bibinfo{person}{Harri Valpola}.} \bibinfo{year}{2017}\natexlab{}.
\newblock \showarticletitle{Mean teachers are better role models:
  Weight-averaged consistency targets improve semi-supervised deep learning
  results}.
\newblock \bibinfo{journal}{\emph{Advances in neural information processing
  systems}}  \bibinfo{volume}{30} (\bibinfo{year}{2017}).
\newblock


\bibitem[Tola et~al\mbox{.}(2012)]%
        {tola2012efficient}
\bibfield{author}{\bibinfo{person}{Engin Tola}, \bibinfo{person}{Christoph
  Strecha}, {and} \bibinfo{person}{Pascal Fua}.}
  \bibinfo{year}{2012}\natexlab{}.
\newblock \showarticletitle{Efficient large-scale multi-view stereo for ultra
  high-resolution image sets}.
\newblock \bibinfo{journal}{\emph{Machine Vision and Applications}}
  \bibinfo{volume}{23}, \bibinfo{number}{5} (\bibinfo{year}{2012}),
  \bibinfo{pages}{903--920}.
\newblock


\bibitem[Wang et~al\mbox{.}(2021)]%
        {wang2021patchmatchnet}
\bibfield{author}{\bibinfo{person}{Fangjinhua Wang}, \bibinfo{person}{Silvano
  Galliani}, \bibinfo{person}{Christoph Vogel}, \bibinfo{person}{Pablo
  Speciale}, {and} \bibinfo{person}{Marc Pollefeys}.}
  \bibinfo{year}{2021}\natexlab{}.
\newblock \showarticletitle{Patchmatchnet: Learned multi-view patchmatch
  stereo}. In \bibinfo{booktitle}{\emph{Proceedings of the IEEE/CVF Conference
  on Computer Vision and Pattern Recognition}}. \bibinfo{pages}{14194--14203}.
\newblock


\bibitem[Wang and Shen(2020)]%
        {wang2020flow}
\bibfield{author}{\bibinfo{person}{Kaixuan Wang} {and} \bibinfo{person}{Shaojie
  Shen}.} \bibinfo{year}{2020}\natexlab{}.
\newblock \showarticletitle{Flow-motion and depth network for monocular stereo
  and beyond}.
\newblock \bibinfo{journal}{\emph{IEEE Robotics and Automation Letters}}
  \bibinfo{volume}{5}, \bibinfo{number}{2} (\bibinfo{year}{2020}),
  \bibinfo{pages}{3307--3314}.
\newblock


\bibitem[Wei et~al\mbox{.}(2021)]%
        {wei2021aa}
\bibfield{author}{\bibinfo{person}{Zizhuang Wei}, \bibinfo{person}{Qingtian
  Zhu}, \bibinfo{person}{Chen Min}, \bibinfo{person}{Yisong Chen}, {and}
  \bibinfo{person}{Guoping Wang}.} \bibinfo{year}{2021}\natexlab{}.
\newblock \showarticletitle{Aa-rmvsnet: Adaptive aggregation recurrent
  multi-view stereo network}. In \bibinfo{booktitle}{\emph{Proceedings of the
  IEEE/CVF International Conference on Computer Vision}}.
  \bibinfo{pages}{6187--6196}.
\newblock


\bibitem[Xiaodong et~al\mbox{.}(2020)]%
        {xiaodong2020cascade}
\bibfield{author}{\bibinfo{person}{Gu Xiaodong}, \bibinfo{person}{Fan Zhiwen},
  \bibinfo{person}{Zhu Siyu}, \bibinfo{person}{Dai Zuozhuo},
  \bibinfo{person}{Tan Feitong}, {and} \bibinfo{person}{Tan Ping}.}
  \bibinfo{year}{2020}\natexlab{}.
\newblock \showarticletitle{Cascade Cost Volume For High-Resolution Multi-View
  Stereo And Stereo Matching}.
\newblock \bibinfo{journal}{\emph{Proceedings of the IEEE conference on
  computer vision and pattern recognition}} (\bibinfo{year}{2020}),
  \bibinfo{pages}{2492--2501}.
\newblock


\bibitem[Xie et~al\mbox{.}(2020)]%
        {xie2020unsupervised}
\bibfield{author}{\bibinfo{person}{Qizhe Xie}, \bibinfo{person}{Zihang Dai},
  \bibinfo{person}{Eduard Hovy}, \bibinfo{person}{Thang Luong}, {and}
  \bibinfo{person}{Quoc Le}.} \bibinfo{year}{2020}\natexlab{}.
\newblock \showarticletitle{Unsupervised data augmentation for consistency
  training}.
\newblock \bibinfo{journal}{\emph{Advances in Neural Information Processing
  Systems}}  \bibinfo{volume}{33} (\bibinfo{year}{2020}),
  \bibinfo{pages}{6256--6268}.
\newblock


\bibitem[Xu et~al\mbox{.}(2021a)]%
        {xu2021self}
\bibfield{author}{\bibinfo{person}{Hongbin Xu}, \bibinfo{person}{Zhipeng Zhou},
  \bibinfo{person}{Yu Qiao}, \bibinfo{person}{Wenxiong Kang}, {and}
  \bibinfo{person}{Qiuxia Wu}.} \bibinfo{year}{2021}\natexlab{a}.
\newblock \showarticletitle{Self-supervised multi-view stereo via effective
  co-segmentation and data-augmentation}. In
  \bibinfo{booktitle}{\emph{Proceedings of the AAAI Conference on Artificial
  Intelligence}}, Vol.~\bibinfo{volume}{2}. \bibinfo{pages}{6}.
\newblock


\bibitem[Xu et~al\mbox{.}(2021b)]%
        {xu2021digging}
\bibfield{author}{\bibinfo{person}{Hongbin Xu}, \bibinfo{person}{Zhipeng Zhou},
  \bibinfo{person}{Yali Wang}, \bibinfo{person}{Wenxiong Kang},
  \bibinfo{person}{Baigui Sun}, \bibinfo{person}{Hao Li}, {and}
  \bibinfo{person}{Yu Qiao}.} \bibinfo{year}{2021}\natexlab{b}.
\newblock \showarticletitle{Digging into Uncertainty in Self-supervised
  Multi-view Stereo}. In \bibinfo{booktitle}{\emph{Proceedings of the IEEE/CVF
  International Conference on Computer Vision}}. \bibinfo{pages}{6078--6087}.
\newblock


\bibitem[Xu and Tao(2020a)]%
        {xu2020learning}
\bibfield{author}{\bibinfo{person}{Qingshan Xu} {and} \bibinfo{person}{Wenbing
  Tao}.} \bibinfo{year}{2020}\natexlab{a}.
\newblock \showarticletitle{Learning inverse depth regression for multi-view
  stereo with correlation cost volume}. In
  \bibinfo{booktitle}{\emph{Proceedings of the AAAI Conference on Artificial
  Intelligence}}, Vol.~\bibinfo{volume}{34}. \bibinfo{pages}{12508--12515}.
\newblock


\bibitem[Xu and Tao(2020b)]%
        {xu2020pvsnet}
\bibfield{author}{\bibinfo{person}{Qingshan Xu} {and} \bibinfo{person}{Wenbing
  Tao}.} \bibinfo{year}{2020}\natexlab{b}.
\newblock \showarticletitle{Pvsnet: Pixelwise visibility-aware multi-view
  stereo network}.
\newblock \bibinfo{journal}{\emph{arXiv preprint arXiv:2007.07714}}
  (\bibinfo{year}{2020}).
\newblock


\bibitem[Yan et~al\mbox{.}(2020)]%
        {yan2020dense}
\bibfield{author}{\bibinfo{person}{Jianfeng Yan}, \bibinfo{person}{Zizhuang
  Wei}, \bibinfo{person}{Hongwei Yi}, \bibinfo{person}{Mingyu Ding},
  \bibinfo{person}{Runze Zhang}, \bibinfo{person}{Yisong Chen},
  \bibinfo{person}{Guoping Wang}, {and} \bibinfo{person}{Yu-Wing Tai}.}
  \bibinfo{year}{2020}\natexlab{}.
\newblock \showarticletitle{Dense hybrid recurrent multi-view stereo net with
  dynamic consistency checking}. In \bibinfo{booktitle}{\emph{European
  Conference on Computer Vision}}. Springer, \bibinfo{pages}{674--689}.
\newblock


\bibitem[Yang et~al\mbox{.}(2020)]%
        {yang2020cost}
\bibfield{author}{\bibinfo{person}{Jiayu Yang}, \bibinfo{person}{Wei Mao},
  \bibinfo{person}{Jose~M Alvarez}, {and} \bibinfo{person}{Miaomiao Liu}.}
  \bibinfo{year}{2020}\natexlab{}.
\newblock \showarticletitle{Cost volume pyramid based depth inference for
  multi-view stereo}. In \bibinfo{booktitle}{\emph{Proceedings of the IEEE/CVF
  Conference on Computer Vision and Pattern Recognition}}.
  \bibinfo{pages}{4877--4886}.
\newblock


\bibitem[Yao et~al\mbox{.}(2018)]%
        {yao2018mvsnet}
\bibfield{author}{\bibinfo{person}{Yao Yao}, \bibinfo{person}{Zixin Luo},
  \bibinfo{person}{Shiwei Li}, \bibinfo{person}{Tian Fang}, {and}
  \bibinfo{person}{Long Quan}.} \bibinfo{year}{2018}\natexlab{}.
\newblock \showarticletitle{Mvsnet: Depth inference for unstructured multi-view
  stereo}. In \bibinfo{booktitle}{\emph{Proceedings of the European Conference
  on Computer Vision (ECCV)}}. \bibinfo{pages}{767--783}.
\newblock


\bibitem[Yao et~al\mbox{.}(2019)]%
        {yao2019recurrent}
\bibfield{author}{\bibinfo{person}{Yao Yao}, \bibinfo{person}{Zixin Luo},
  \bibinfo{person}{Shiwei Li}, \bibinfo{person}{Tianwei Shen},
  \bibinfo{person}{Tian Fang}, {and} \bibinfo{person}{Long Quan}.}
  \bibinfo{year}{2019}\natexlab{}.
\newblock \showarticletitle{Recurrent mvsnet for high-resolution multi-view
  stereo depth inference}. In \bibinfo{booktitle}{\emph{Proceedings of the
  IEEE/CVF Conference on Computer Vision and Pattern Recognition}}.
  \bibinfo{pages}{5525--5534}.
\newblock


\bibitem[Yao et~al\mbox{.}(2020)]%
        {yao2020blendedmvs}
\bibfield{author}{\bibinfo{person}{Yao Yao}, \bibinfo{person}{Zixin Luo},
  \bibinfo{person}{Shiwei Li}, \bibinfo{person}{Jingyang Zhang},
  \bibinfo{person}{Yufan Ren}, \bibinfo{person}{Lei Zhou},
  \bibinfo{person}{Tian Fang}, {and} \bibinfo{person}{Long Quan}.}
  \bibinfo{year}{2020}\natexlab{}.
\newblock \showarticletitle{Blendedmvs: A large-scale dataset for generalized
  multi-view stereo networks}. In \bibinfo{booktitle}{\emph{Proceedings of the
  IEEE/CVF Conference on Computer Vision and Pattern Recognition}}.
  \bibinfo{pages}{1790--1799}.
\newblock


\bibitem[Yi et~al\mbox{.}(2020)]%
        {yi2020pyramid}
\bibfield{author}{\bibinfo{person}{Hongwei Yi}, \bibinfo{person}{Zizhuang Wei},
  \bibinfo{person}{Mingyu Ding}, \bibinfo{person}{Runze Zhang},
  \bibinfo{person}{Yisong Chen}, \bibinfo{person}{Guoping Wang}, {and}
  \bibinfo{person}{Yu-Wing Tai}.} \bibinfo{year}{2020}\natexlab{}.
\newblock \showarticletitle{Pyramid multi-view stereo net with self-adaptive
  view aggregation}. In \bibinfo{booktitle}{\emph{European Conference on
  Computer Vision}}. Springer, \bibinfo{pages}{766--782}.
\newblock


\bibitem[Yu and Gao(2020)]%
        {yu2020fast}
\bibfield{author}{\bibinfo{person}{Zehao Yu} {and} \bibinfo{person}{Shenghua
  Gao}.} \bibinfo{year}{2020}\natexlab{}.
\newblock \showarticletitle{Fast-mvsnet: Sparse-to-dense multi-view stereo with
  learned propagation and gauss-newton refinement}. In
  \bibinfo{booktitle}{\emph{Proceedings of the IEEE/CVF Conference on Computer
  Vision and Pattern Recognition}}. \bibinfo{pages}{1949--1958}.
\newblock


\bibitem[Zhang et~al\mbox{.}(2020)]%
        {zhang2020label}
\bibfield{author}{\bibinfo{person}{Yabin Zhang}, \bibinfo{person}{Bin Deng},
  \bibinfo{person}{Kui Jia}, {and} \bibinfo{person}{Lei Zhang}.}
  \bibinfo{year}{2020}\natexlab{}.
\newblock \showarticletitle{Label propagation with augmented anchors: A simple
  semi-supervised learning baseline for unsupervised domain adaptation}. In
  \bibinfo{booktitle}{\emph{European Conference on Computer Vision}}. Springer,
  \bibinfo{pages}{781--797}.
\newblock


\bibitem[Zhu et~al\mbox{.}(2021)]%
        {zhu2021multi}
\bibfield{author}{\bibinfo{person}{Jie Zhu}, \bibinfo{person}{Bo Peng},
  \bibinfo{person}{Wanqing Li}, \bibinfo{person}{Haifeng Shen},
  \bibinfo{person}{Zhe Zhang}, {and} \bibinfo{person}{Jianjun Lei}.}
  \bibinfo{year}{2021}\natexlab{}.
\newblock \showarticletitle{Multi-View Stereo with Transformer}.
\newblock \bibinfo{journal}{\emph{arXiv preprint arXiv:2112.00336}}
  (\bibinfo{year}{2021}).
\newblock


\end{thebibliography}

\end{document}